\documentclass[11pt, dina4,  amssymb]{article}
\pdfoutput=1
\usepackage[margin=1in]{geometry}
\usepackage[normalem]{ulem}

\usepackage[bookmarks=true, colorlinks, citecolor=blue, linkcolor=black]{hyperref}
\usepackage{graphicx}
\usepackage{epstopdf}
\usepackage{authblk}
\usepackage{amsmath, amsthm, amssymb, graphicx, multirow, subfigure, float, ulem, threeparttable}
\usepackage[bookmarks=true, colorlinks, citecolor=blue, linkcolor=black]{hyperref}
\usepackage{geometry, adjustbox}
\usepackage{caption}
\usepackage{CJKutf8}
\begin{document}
\begin{CJK}{UTF8}{gbsn}
\title{ClusTop: An unsupervised and integrated text clustering and topic extraction framework}
\author[a,1]{Zhongtao Chen}
\author[a,1]{Chenghu Mi}
\author[b,2]{Siwei Duo}
\author[a]{Jingfei He}
\author[a]{Yatong Zhou}
\affil[a]{School of Electronics and Information Engineering, Hebei University of Technology}
\affil[b]{Tianjin Huizhi Xingyuan Information Technology Co.,Ltd.}
\footnotetext[1]{The first two authors contributed equally (listed in alphabetical order). }
\footnotetext[2]{Corresponding author: siweiduo@cispirit.com}
\date{}
\maketitle

\begin{abstract} 
  Text clustering and topic extraction are two important tasks in text mining. 
  Usually, these two tasks are performed separately. 
  For topic extraction to facilitate clustering, we can first project texts into a topic space and then perform a clustering algorithm to obtain clusters. 
  To promote topic extraction by clustering, we can first obtain clusters with a clustering algorithm and then extract cluster-specific topics. 
  However, this naive strategy ignores the fact that text clustering and topic extraction are strongly correlated and follow a chicken-and-egg relationship. 
  Performing them separately fails to make them mutually benefit each other to achieve the best overall performance.
  In this paper, we propose an unsupervised text clustering and topic extraction framework (ClusTop) which integrates text clustering and topic extraction into a unified framework and can achieve high-quality clustering result and extract topics from each cluster simultaneously.
  Our framework includes four components: enhanced language model training, dimensionality reduction, clustering and topic extraction, where the enhanced language model can be viewed as a bridge between clustering and topic extraction.
  On one hand, it provides text embeddings with a strong cluster structure which facilitates effective text clustering; 
  on the other hand, it pays high attention on the topic related words for topic extraction because of its self-attention architecture.
  Moreover, the training of enhanced language model is unsupervised.
  Experiments on two datasets demonstrate the effectiveness of our framework and provide benchmarks for different model combinations in this framework.

\end{abstract}

\section{Introduction}
  With the rapid development of information technology, huge amounts of data are created every day.
  Automatic text clustering and topic extraction are intuitively important, as they facilitate many web applications such as question-answering systems \cite{Zhao2008}, public opinion monitoring \cite{Chen2019}, and personalized recommendations \cite{Shepitsen2008}. 
  The two main components of text clustering are text representation and clustering. 
  Existing text clustering methods mainly focus on optimizing text representation models and clustering algorithms to achieve better clustering results. \par
  Text representation models convert the target texts into vector representations by capturing the semantic information so that it can be processed by the clustering algorithm.
  Recently, pre-trained language models (such as BERT \cite{Devlin2018}) have shown great performance in generating contextual word and sentence embeddings.
  However, BERT-based models induce an anisotropic semantic space of sentences \cite{Li2020}, which harms their performance in tasks related to semantic similarity, such as text clustering.
  To alleviate this problem, BERT-flow \cite{Li2020} and BERT-whitening \cite{Su2020} apply post-processing by mapping the obtained embeddings to an isotropic distribution.
  SimCSE \cite{Gao2021}, ConSERT \cite{Yan2021}, ESimCSE \cite{Wu2022} and DiffCSE \cite{Chuang2022} presented contrastive frameworks to eliminate anisotropy in sentence embeddings.
 Although the above methods have been shown effective in the task of sentence semantic similarity, they are not good enough to be directly used for text clustering because sentences in the same cluster but with different meanings are overly distinguished.
 The distance between sentences in the same cluster may event exceed the distance between sentences in different clusters, and the overall distribution of sentences cannot show a clear cluster structure, which is not conducive to the application of subsequent clustering algorithm.
 For this reason, we propose an unsupervised approach to obtain an enhanced language model which can generate sentence embeddings with a well-seperated and balanced cluster structure, making it more suitable for clustering task.
 Specifically, the enhanced language model is obtained by fine-tuning a pre-trained language model on a classification task with pseudo-labeled data.
 The pseudo-labeled data is obtained through a clustering-based pseudo-labeling approach by first using the pre-trained language model to obtain the sentence embeddings, then applying UMAP (Uniform Manifold Approximation and Projection) \cite{McInnes2018} for dimensionality reduction and a density-based clustering algorithm, such as HDBSCAN (Hierarchical Density-Based Spatial Clustering of Applications with Noise) \cite{McInnes2017}, for selecting non-noisy data in high-density region.\par
 
  Since the sentence embeddings generated by the enhanced language model are high dimensional, Euclidean distance based clustering algorithms are ineffective because of the curse of dimensionality.
  A straight-forward approach is to reduce the dimensionality of embeddings. 
  The existing dimensionality reduction techniques can be divided into two categories: linear \cite{Jolliffe2016, Sarwar2000, Izenman2013} and nonlinear \cite{2-12, Bach2002, Baudat2000, VanderMaaten2008},
  among which PCA \cite{Jolliffe2016} and t-SNE \cite{VanderMaaten2008} are the two well-known methods.
  While, UMAP \cite{McInnes2018} has shown to capture more global structure than t-SNE and has superior runtime performance.
  In addition, UMAP has a solid theoretical foundations as a manifold approximation and projection.
  Since the number of clusters is unknown and the texts to be clustered might include noice points, to avoid assign those points to any unrelated cluster, it is better to choose algorithms, such as DBSCAN \cite{Ester1996} and HDBSCAN \cite{McInnes2017}.
  In our experiments, we also include the well-known partition-based clustering algorithm k-Means as a comparison by giving it a reasonable cluster number.
\par

  From the clustered texts, the traditional keywords extraction methods TF-IDF \cite{Ramos2003,Trstenjak2014} and TextRank \cite{Mihalcea2004} can be used to extract the cluster-specific topics. 
  TF-IDF is simple to implement but it doesn't capture semantics and relies heavily on corpus and the value of IDF.
  TextRank has high computational complexity and omits keywords that have a lower chance of appearing, although they are significant in context.
  Regardless of the clustering result, the traditional topic model LDA \cite{Blei2003} can be used independently to extract topics but it requires the number of topics to be known ahead and is limited by its assumptions of the data.
  Taking advantage of the semantic information learned by pre-trained language model, keyBERT \cite{2019Sharma} leverages embeddings from BERT to find keywords that are most similar to a document.
  However, due to the maximum length accepted by BERT (no more than 512 tokens), only part of the sentences in a given cluster can be selected for concatenation until the total length is reached, which will undoubtedly result in a loss of accuracy. \par
  
  In order to get the keywords to describe the main topic of each cluster, we propose an attention based topic extraction method. 
  The core of our method is to maximize the potential of the enhanced language model which has been trained with strong classification ability.
  On the one hand, it can provide sentence embeddings that contain strong categorical features,  making clustering easier; on the other hand, the most important component of language models, self-attention mechanisms, can tell us which tokens in the sentence are playing an important role, thus contributing to the task of topic extraction.
  Based on an intuitive perception, the more important a token is in a sentence, the higher the attention paid to it by the other tokens, 
  our method uses the sum of the attention weights of all tokens in the sentence for a given token to measure its importance in the sentence, thus obtaining the topic words for each cluster.
  It is worth noting that the enhanced language model naturally integrating the clustering and topic extraction process into one.\par
  Our main contributions can be summarized as follows:\par
  1) We propose an unsupervised and integrated text clustering and topic extraction framework ClusTop, which consists of enhanced language model training, dimensionality reduction, clustering, and attention based topic extraction. 
  Experiments on two datasets demonstrate the effectiveness of our framework.
  Benchmarks have been provided for different model combinations.\par
  2) We propose an unsupervised approach to obtain an enhanced language model which can generate sentence representations with a distinct and balanced cluster shape, making it more friendly for the next step of clustering.\par
  3) We propose an attention based topic extraction method for obtaining keywords for each cluster. 
  The obtained keywords are more close to the cluster semantics, as the potential of the self-attention mechanism is mined out.\par
  The paper is organized as follows: Section II discusses the related works. In Section III, the proposed text clustering and topic extraction framework is described in detail. Section IV presents the experimental results, and conclusions are given in the following section.

\section{Related Works}
Text clustering and topic extraction are two correlated tasks in the field of text mining. 
First clustering the text, then extracting keywords for each category has become one of the most representative frameworks for topic extraction.
To the best of our knowledge, the most advanced results based on this framework are Top2Vec \cite{Angelov2020} and the recently proposed BERTopic \cite{Grootendorst2022}. 
These two works were originally developed for topic extraction, but both were carried out on the basis of clustering.
Specifically, they all first adopted an embedding, dimensionality reduction, and clustering process to cluster the text, and then extracted the keywords of each cluster as a description of its topic.
In the text embedding step, Top2Vec is based on doc2vec  \cite{Le2014}, while BERTopic is based on a pre-trained language model, they both use UMAP and HDBSCAN for dimensionality reduction and clustering, respectively.
The main difference between these two methods lies in how to extract keywords from each cluster.
To obtain the topic words for each cluster, Top2Vec takes the centroid of the cluster as the topic vector, and believes that the closer the word embedding is to the topic vector, the better it can represent the topic. 
However, there is a clear incompatibility between clustering and topic word selection - clustering is based on the density of the data distribution, while word selection is based on the distance from the cluster centroid.
For clusters that are strip-shaped or irregularly shaped, this method is prone to mistakenly select words from other clusters.
Reducing the radius of word selection is the most intuitive solution to the aforementioned problem, but it will miss a lot of candidate words.
BERTopic proposed a class-based variation of TF-IDF (c-TF-IDF) to extract topic words from each cluster, which overcomes the centroid-related issues.
However, due to its TF-IDF nature, it will still be deeply affected by word frequency and number of clusters.
It is worth noting that neither of the two methods take advantage of the distributed representations of pre-trained language models, which can better capture semantics compared to frequency-based statistical methods.
\par
Our framework is fundamentally different from the aforementioned two methods in two aspects: text embedding and topic extraction.
In terms of text embedding, our proposed method doesn't directly apply a pre-trained language model, instead, we replace it with a cluster-enhanced language model (i.e., a language model with strong category representation capability) and the enhanced language model is trained unsupervisedly.
Specifically, embeddings obtained from the original pre-trained language model are more evenly distributed in the vector space with less obvious cluster structure. 
Directly applying HDBSCAN for clustering will result in a large proportion of noise points, which leads to a low recall rate of the results. 
In contrast, text embeddings obtained from the enhanced language model have a distinct cluster structure, with points in the cluster being more compact, distances between different clusters being larger and boundaries being clearer. 
This undoubtedly makes subsequent HDBSCAN clustering more beneficial, leading to a reduction in the number of noise points and an increase in the recall rate.
In terms of topic extraction, our method maximizes the advantage of the enhanced language model by directly applying its self-attention mechanism to the topic extraction task.
Inspired by the fact that the enhanced language model yields good text clustering results, it indicates that the enhanced language model already has a strong ability to distinguish categories.
From the sentence-level perspective, the enhanced language model provides embeddings with a more distinct cluster shape which benefits clustering; 
from the token-level perspective, it shifts its attention to those tokens which are most relevant to the cluster topics, thus naturally providing the most fitting topic words for each cluster.
Compared to the aforementioned methods, our method uses the enhanced language model to glue together clustering and topic extraction, maximizing the semantic representation capability of the language model.

\section{Framework}
\begin{figure}[htbp]
	\centering
	\resizebox{\textwidth}{!}{
	\includegraphics[width=16cm]{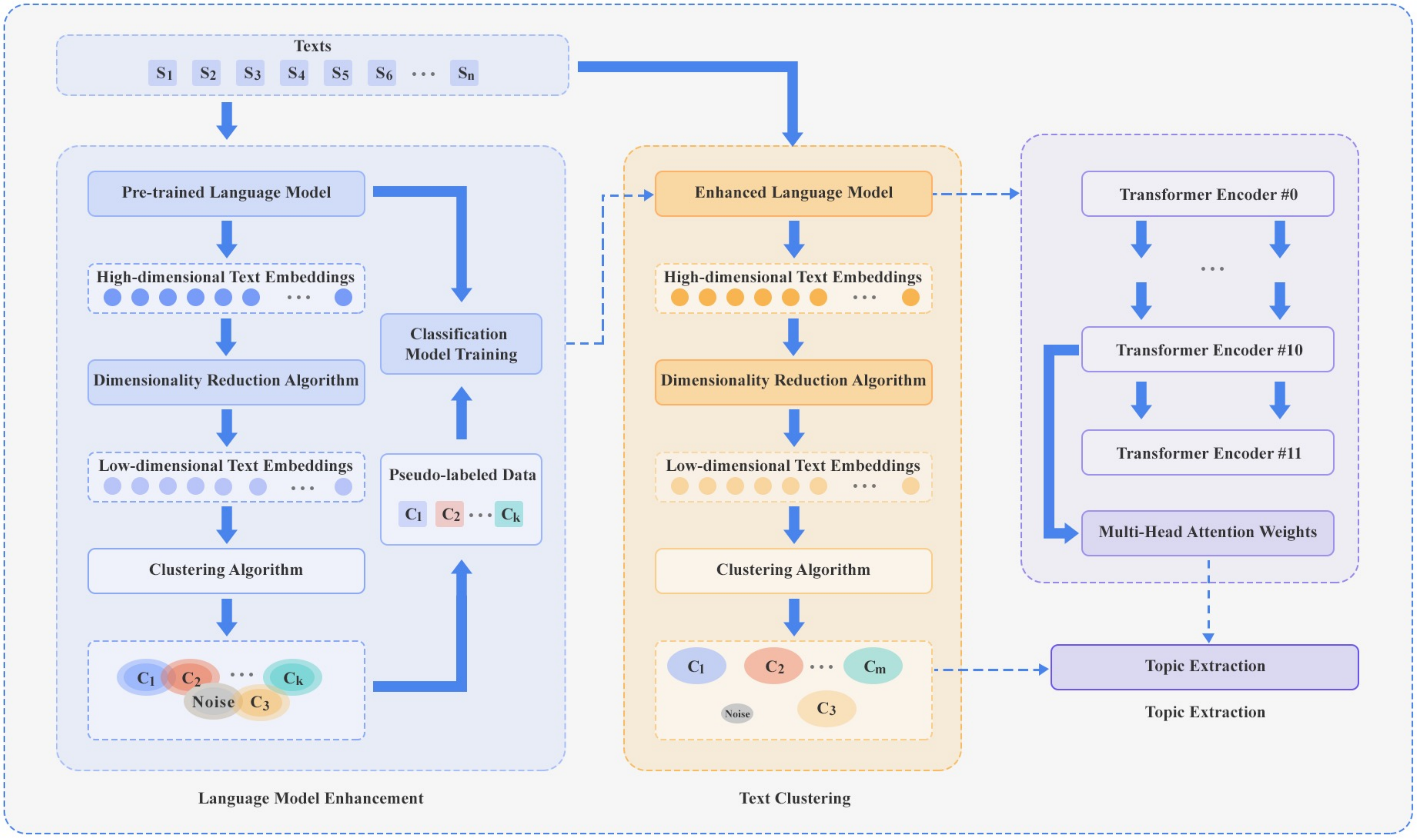}
	}
	\captionsetup{font=scriptsize}\caption{The first part of the framework is the acquisition of the enhanced language model with strong cluster features; the second part includes text representation based on the enhanced language model, dimensionality reduction, and clustering; and the third part is the attention based topic extraction.}
\label{Figure-1}
\end{figure}

Our text clustering and topic extraction framework is mainly divided into two parts: an enhanced language model based text clustering and an attention based topic extraction, as shown in Figure  \ref{Figure-1}.
First, an enhanced language model needs to be obtained for text representation.
The enhanced language model can be treated as a promotion of the original pre-trained language model, which can enlarge the distance between clusters and making the points within the cluster more compact, thus making the generated text embeddings more conducive to clustering task.
Due to the high dimensionality of the obtained text embeddings, dimensionality reduction is processed before clustering.
Finally, an attention based topic extraction method is applied to extract topics from each cluster.
In the following, we will describe our text clustering and topic extraction framework in detail,
and the focus will be on how to obtain the enhanced language model and how to extract topics based on the attention mechanism.

\subsection{An enhanced language model based text clustering}
The transformer-based pre-trained language models (without fine-tuning) have certain semantic representation capabilities, but they are not good enough to be directly used for text clustering because the distribution of text embeddings do not have an obvious cluster structure, thus leads to a low accuracy in clustering results.
Specifically, due to the lack of a clear cluster structure, partition-based clustering methods will misclassify points due to the irregular boundaries between different clusters; density-based clustering methods will result in a large number of noise points, which should belong to certain clusters.
Inspired by the fine-tuning paradigm, to address the above issues, 
we train the language model on a text classification task to learn the cluster-level semantic, making the enhanced language model more sensitive to the cluster features of sentences.
Applying the cluster features-enhanced language model for text embedding results in a text representation space with a clear cluster structure, which is more conducive for clustering.
This section includes three steps: training of the cluster features-enhanced language model, dimensionality reduction, and clustering. 
The training of the enhanced language model will be described in detail, which is also one of the main contributions of our work.
\par

\noindent
\textbf{Training of the enhanced language model}

Inspired by the pretrain-then-finetune paradigm, we train the language model to learn semantic information at the cluster level through training a classification task, 
so as to adjust the distribution of the original text representation to make it have a stronger cluster characteristic, with points within the same cluster being more compact and boundaries between different clusters being more distinct.
To obtain the labeled data for training the classification model, we propose an unsupervised framework which can be divided into four steps: text representation, dimensionality reduction, clustering and classification model training.
\par

Step 1. Text representation

For the text representation, we use the unsupervised SimCSE \cite{Gao2021} pre-trained language model.
Directly using the sentence embeddings obtained from BERT for tasks related to semantic similarity has poor performance due to the anisotropic in sentence embeddings.
BERT-flow maps the sentence embedding distribution into an isotropic Gaussian distribution through normalizing flows; 
BERT-whitening transforms the sentence embeddings to follow a standard normal distribution，but is unable to deal with nonlinear dependencies between features.
Based on the idea of contrastive learning by pulling close similar samples and pushing away dissimilar samples, SimCSE greatly improves state-of-the-art sentence embeddings on semantic textual similarity tasks.
For this reason, we simply take the unsupervised SimCSE for sentence embedding.

Step 2. Dimensionality reduction

After being embedded by SimCSE, sentences are represented as 768-dimensional vectors. 
Since most clustering algorithms cannot handle high dimensional vectors well and high dimensional vectors usually require a high computational cost, it is necessary to reduce the dimension of text embeddings before clustering.
Dimensionality reduction not only can facilitate data processing, but also can help to visualize the distribution of text embeddings.
In this step, we apply a manifold-based dimensionality reduction algorithm - UMAP, which is developed based on Riemannian geometry and algebraic topology and is grounded in a strong mathematical foundation.
Compared to t-SNE, UMAP not only preserves more global structures, but also has an increased speed.
The two most commonly used parameters in UMAP are n-neighbors and min-dist, adjusting these parameters can effectively control the balance between the local structure and the global structure in the dimensionality reduction results.
The n-neighbors parameter controls how UMAP balances local versus global structure by constraining the size of the local neighborhood. 
Small values will encourage UMAP to focus more on local structure, while large values will encourage UMAP to favor representing the global structure while losing fine details.
The min-dist parameter represents the minimum distance between points in the low-dimensional space, and this parameter controls how tightly UMAP is allowed to cluster points together.
When the value is set lower, the points in the low-dimensional embedding space are distributed with finer topological structure; while at a higher value setting, UMAP will pack the points together more loosely and focus more on preserving the broad topological structure.
\par


Step 3. Clustering

Due to the irregular shapes of data distributions in the low-dimensional space, uneven density distributions, and the existing of noise points, we adopt the density-based clustering algorithm HDBSCAN.
Rather than finding clusters with a particular shape, HDBSCAN looks for regions that are denser than the surrounding space, it is noice aware and can find clusters even with some arbitrary shape.
HDBSCAN has two important parameters, metric and min-cluster-size.
The metric parameter is used to measure the distance between vectors, we use Euclidean metric in all experiments in this paper.
The min-cluster-size parameter represents the minimum cluster size, which will also influence the number of topics. The larger the value, the fewer the number of clusters, and vice versa.
In our experiments, the min-cluster-size will be adjusted according to the characteristics of different datasets.
\par

Unlike the dimensionality reduction and clustering steps in the main framework which can be any combination of dimensionality reduction and clustering algorithms, we particularly choose UMAP combined with HDBSCAN in this step to guarantee the high quality of the pseudo-labeled data.
This combination takes advantage of UMAP's ability to preserve the global and topological structure of the data, and because HDBSCAN can cluster data with different densities, the non-noise points in the higher density areas can be naturally selected as the pseudo-labeled data.
After the text embeddings are dimensionality reduced and clustered, each non-noise data is labeled and can be used to train a classification model.
\par

Step 4. Classification model training

The cluster feature enhanced language model can be obtained by training a classification model, which consists of the original pre-trained language model with an additional fully-connected classifier on top.
The enhanced language model can be obtained by first training the classification model on the pseudo-labeled data, and then removing the fully-connected layer from the trained classification model.
Compared to the original pre-trained language model, the enhanced language model can obtain more cluster semantic information through the training of the classification task, making it able to provide sentence embedding with more obvious cluster structures.

\noindent
\textbf{Text clustering based on the enhanced language model}

Since the enhanced language model has better ability to represent the cluster semantics, it is more conducive to dimensionality reduction and clustering and less strict to the choice of algorithms.
The dimensionality reduction and clustering algorithms in the our main framework are not limited to the combination of UMAP and HDBSCAN, and can be replaced by different dimensionality reduction and clustering algorithms.
In fact, the key point of this part in our framework lies in how to obtain a language model with more obvious cluster structure.
With the power of the enhanced language model, the obtained text embeddings will be more tolerant to different combinations of subsequent dimensionality reduction and clustering algorithms.
In the experimental section, we will compare different combinations of dimensionality reduction and clustering algorithms.
\par

\subsection{An attention based topic extraction}
Dimension reduction and clustering based on the enhanced language model can achieve good text clustering results, which indicates that the embeddings obtained from the enhanced language model have already contained the topic semantic information, and these topic semantics are embodied in some attention layers of the language model.
Specifically, some of the attention layers are able to pay more attention to the tokens that are related to the topic semantics, giving them a higher attention weight.
This also aligns with the intuitive assumption that the more a given word can represent the meaning of a sentence, the higher the attention weight of the other words in the sentence for that given word.
Inspired by this, we propose a topic extraction method based on the attention mechanism, using the intermediate product of the language model - the attention weight of each token to measure the importance of the token in the sentence.
For a given sentence, our method first sums up the attention weights from all tokens to the given token in the 10th layer of the Transformer as the importance score of that given token; then select the token with the highest score to be the keyword of the given sentence; finally, for a given cluster, sorts all the keywords according to their frequency, and then selects the top-$K$ as the topic words.
Detail steps are described as follows.

\par
Step 1. Sentence key token extraction

For a given sentence, we assume that the values of tokens in the sentence are $\{\textbf{v}_1, \textbf{v}_2, ..., \textbf{v}_n\}$, the matrix $\{\alpha_{i,j}\}_{i,j=1}^n$ represents the attention weights matrix obtained from the self-attention layer, where $\alpha_{i,j}$ represents the attention score from the $i$-th token to the $j$-th token，taking the weighted summation, we can obtain the representation of the $i$-th token as follows：
$$ \textbf{e}_i = \sum_{j=1}^{n} \alpha_{i,j} \textbf{v}_j, \quad i = 1,2,...,n.$$
The vector representation of the sentence can be obtained by taking the average of the word vectors obtained above：
$$ \textbf{e} = \frac{1}{n}\sum_{i=1}^{n} \textbf{e}_i 
              = \frac{1}{n}\sum_{i=1}^{n} \sum_{j=1}^{n} \alpha_{i,j} \textbf{v}_j
              = \frac{1}{n}\sum_{j=1}^{n} (\sum_{i=1}^{n} \alpha_{i,j}) \textbf{v}_j
              = \frac{1}{n}\sum_{j=1}^{n} \beta_{j} \textbf{v}_j
              ,$$
where $\beta_{j}=\sum_{i=1}^{n} \alpha_{i,j}$ represents the sum of the attention weights from all tokens in the sentence to the $j$-th token.
Intuitively, if a token can serve as the key of its sentence, it will receive more attention from all the tokens in the sentence.
In other words, all the tokens in the sentence will give large attention to the key token, the value of $\beta_{j}$ according to the key token (the $j$-th token) will be large.
It needs to notice that the above analysis is not mathematically rigorous, and is only intended to provide some analytical support for our intuition.
\par

\begin{figure}[htbp]
	\centering
	\includegraphics[width=14cm]{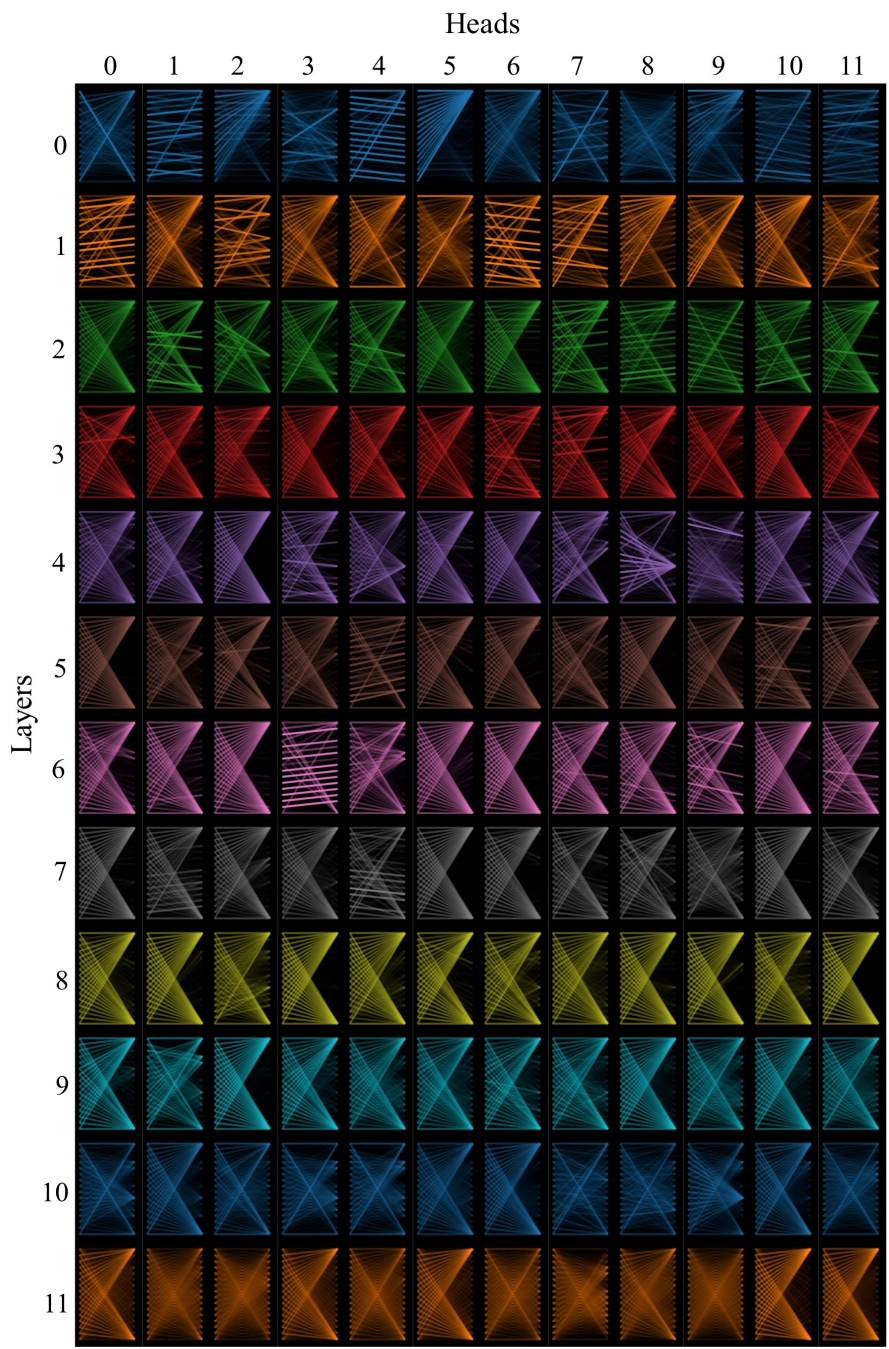}
	\captionsetup{font=scriptsize}\caption{The figure shows the 12-layer Transformer attention weights by the enhanced-SimCSE-chinese-roBERTa model of sentence '多家银行封杀信用卡支付宝交易(A lot of banks have blocked transactions with credit card in Alipay)', where the columns represent the Transformer layers and the rows represent the attention heads.}
\label{Figure-2}
\end{figure}

We can also visualize the above analysis by the Attention map, so as to understand the relatedness between tokens, as well as the importance of each token in the sentence.
Figure \ref{Figure-2} shows the attention map for the sentence "多家银行封杀信用卡支付宝交易 (A lot of banks have blocked transactions with credit card in Alipay)".
The thickness of the lines represent the attention weight from the left token to the right, the thicker the line, the greater the attention weight.
From the results of the above qualitative analysis, the 10th layer of the Transformer has learned the cluster semantics, and can give greater attention to the tokens related to the cluster topic.
However, since the above result only present a visualization of a single sentence, which do not represent the overall situation. 
Therefore, we will present statistical analysis in the following to compare the sensitivity of different Transformer layers to the cluster semantics.
\par

Table \ref{Table-0} shows various statistical metrics of attention weights in different layers of Transformers base on the enhanced language model.
We select coefficient of variation (Cov.), kurtosis (Kurt.) and relative range (R.R.) as statistical metrics, which reflect the characteristics of whether there is a focus point in the data from different angles. 
The purpose is to select the appropriate Transformer layer that carries more cluster semantics such that it can provide keywords of each sentence.
The coefficient of variation is a measure of relative variability of the data. 
It is calculated by dividing the standard deviation by the mean. 
A smaller value indicates that the data points are closer to the mean, while a larger value indicates that the data points are more spread out.
Kurtosis is used to measure the degree of peakedness of a distribution.
It measures how much of the data is concentrated around the mean compared to the tails of the distribution and reveals the relative levels of extreme values within a dataset.
The higher the kurtosis value, the more peaked the probability distribution is, and the lower the value, the flatter the distribution.
The range is the difference between the largest and smallest values in the data. 
The relative range is the ratio of the range to the average of the data and can be used to indicate the degree of dispersion.

\begin{table}[htbp]\footnotesize
	\centering
	\resizebox{\textwidth}{!}{
	\begin{tabular}{lcccccccccccc}
    \hline
    \textbf{Layer}  & \textbf{0} & \textbf{1} & \textbf{2} & \textbf{3} & \textbf{4} & \textbf{5} & \textbf{6} & \textbf{7} & \textbf{8} & \textbf{9} & \textbf{10} & \textbf{11} \\ \hline
    \textbf{Cov.} & 0.1767 & 0.2422 & 0.2723 & 0.3001 & 0.5308 & 0.4726 & 0.4228 & 0.6646 & 0.7258 & 0.8766 & \underline{1.2354} & 0.2855 \\
    \textbf{Kurt.} & 0.4379 & 0.5333 & 0.6412 & 0.4578 & 2.7852 & 0.1905 & 1.8723 & 3.9464 & 4.2924 & 5.6261 & \underline{8.2876} & 0.6091 \\
    \textbf{R.R.} & 0.6737 & 0.9415 & 1.0570 & 1.1436 & 2.1683 & 1.9643 & 1.6809 & 2.7304 & 2.9495 & 3.6941 & \underline{5.4171} & 1.0846 \\
    \hline
   \end{tabular}
   }	
	\captionsetup{font=scriptsize}\caption{Statistical metrics of the sum of attention weights in different layers. Cov.: the average of the Coefficients of Variation of all sentences; Kurt.: the average of the Kurtosis of all sentences; R.R.: the average of the Relative Range of all sentences. The highest values are underlined of each statistical metric.}
\label{Table-0}
\end{table}
The results in Table \ref{Table-0} suggest that the sums of attention weights corresponding to the 10th layer of the Transformer has higher coefficient of dispersion, peakedness and range than other layers, which indicates that the data has obvious peaks, and these peaks correspond to the keywords of the sentence.
Compared with other layers, the attention weights corresponding to the 10th layer of the Transformer have a higher Cov., Kurt. and R.R., which indicates that the data has obvious peaks, and these peaks correspond to the key tokens of the sentence.
Based on the above qualitative and quantitative analysis, we select the attention weights from the 10th layer of the Transformer to determine the keywords of the sentence.

Step 2. Sentence keywords extraction

In this step, key tokens obtained from Step 1 are mapped to their corresponding words to obtain the keywords of each sentence.
First, we use THULAC for sentence tokenization. 
Then, based on the tokenized sentences, map the key tokens obtained in Step 1 to their corresponding words to get a list of keywords.
For example, for the sentence "楼上下水管道漏水，滴到楼下家里(The water pipe upstairs is leaking and dripping into the downstairs house)", the token with the highest score in Step 1 is "漏(leaking)", the tokenized sentence is "楼上/下水/管道/漏水/，/滴/到/楼下/家里/", mapping "漏" to "漏水", finally we get the keywords of this sentence.

Step 3. Cluster topic words extraction

In this step, we will extract the topic words for each cluster. 
For a given cluster, first we take the union of keywords from all sentences to construct a keyword set for this cluster. 
Then, sort the keywords by word frequency from high to low. 
Finally, take the top $K$ keywords in the list to be the topic words of this cluster.

\section{Experiment}

\subsection{Experiment Setup}

\noindent
\textbf{Datasets}

Two datasets were used to validate the proposed framework: the THUCNews dataset \cite{Sun2016} provided by the Natural Language Processing Laboratory of Tsinghua University and the real-world social governance dataset collected by a local grid member. 
The THUCNews dataset is generated by filtering the historical data of Sina News RSS subscription channels from 2005 to 2011. It contains 740,000 news documents (2.19 GB) and is divided into 14 categories: finance, lottery, real estate, stock, home furnishings, education, technology, society, fashion, political science, sports, constellation, game, and entertainment. 
We select 1000 headline data from each category to form a dataset with a total of 14000. 
The social governance dataset consists of social governance events about living environment, transportation, public facilities, and neighborhood disputes collected by community grid members, with a total of 16,870 events.

\noindent
\textbf{Choice of parameters}

During the experiment, we comprehensively consider the size of the dataset and the number of categories when adjusting n-neighbors in UMAP and t-SNE, min-cluster-size in DBSCAN and HDBSCAN, and eps in DBSCAN. 
The size of the dataset and the number of categories are both considered in this paper while adjusting n-neighbors in UMAP and t-SNE, min-cluster-size in DBSCAN and HDBSCAN, and eps in DBSCAN. 
We traversed the parameters n-neighbors and min-cluster-size between 100-200 with a step of 10, and the parameter eps between 0-4 with a step of 0.1 on the THUCNews dataset, crossing combinations to select the parameter combination with the best NMI.
Considering the complexity and variety of topics of the social governance dataset, we set min-cluster-size=20 and n-neighbors=20 to ensure that the algorithm can extract finer-grained category information.

\subsection{Text Clustering}

The baseline models for comparison in the experiments are LDA \cite{Blei2003}, FastText \cite{Joulin2017} and Top2Vec \cite{Angelov2020}.
\renewcommand{\thefootnote}{\arabic{footnote}}
Based on the proposed framework, we choose Bert-base-Chinese\footnotemark[1]\footnotetext[1]{\url{https://huggingface.co/bert-base-chinese/tree/main}} (BERT), SimCSE\footnotemark[2]\footnotetext[2]{\url{https://huggingface.co/cyclone/SimCSE-chinese-roberta-wwm-ext/tree/main}}, and their enhanced versions (BERT-enhanced and SimCSE-enhanced) as text embedding models; UMAP, t-SNE, and PCA as dimensionality reduction algorithms; K-Means, DBSCAN, and HDBSCAN as clustering algorithms and the effects of cross-combination of different models and algorithms are compared. 
In order to comprehensively evaluate the performance of the proposed framework, we use external evaluation metrics: Adjusted Rand Index (ARI) \cite{HuBERT1985}, Adjusted Mutual Information (AMI) \cite{Vinh2009}, Purity and F1-measure; internal evaluation metrics: Silhouette Coefficient (SC), Kalinski-Harabaz Index (CHI) and Davies-Bouldin Index (DBI).
\par
Table \ref{Table-1} shows the clustering results of different methods on THUCNews dataset. 
We choose LDA, FastText, and Top2Vec as baselines, and compare different combinations of language models (i.e., original and enhanced language models), dimensionality reduction algorithms, and clustering algorithms. 
Compared to the baseline models, different combinations based on the language model perform better on various evaluation metrics. 
In the case of using the same dimensionality reduction algorithm and clustering algorithm, the enhanced language model can achieve a better clustering result than the original language model.
For example, BERT-enhanced+PCA+HDBSCAN improves the ARI dramatically by +26.68\% absolute compared to BERT+PCA+HDBSCAN, which further verifies that the text embeddings obtained by the enhanced language model have stronger clustering properties. 
When using t-SNE or UMAP to perform dimensionality reduction, the enhanced language model does not perform significantly better than the original language model, which is due to the ability of t-SNE and UMAP to preserve the global and local structure of the data. 
Therefore, even if the text embedding obtained by the original language model does not have obvious clustering properties, it can still obtain acceptable clustering results because the spatial distribution characteristics are preserved.\par

Table \ref{Table-1} also shows that BERT performs better than SimCSE before the language model is enhanced. For example, BERT+t-SNE+KMeans and BERT+UMAP+HDBSCAN have achieved the best results in various external evaluation metrics, and BERT+UMAP+HDBSCAN is also the combination used by BERTopic. BERT+UMAP+KMeans performs best in various internal evaluation metrics and is slightly better than SimCSE+UMAP+KMeans. After the language model is enhanced, BERT-enhanced+(t-SNE/UMAP)+(KMeans/DBSCAN/HDBSCAN) performs better on external evaluation metrics, and SimCSE-enhanced+UMAP+HDBSCAN achieves the best performance on internal evaluation metrics.\par

\begin{figure}[htbp]
	\centering
	\subfigure[BERT+UMAP+HDBSCAN]{
		\includegraphics[scale=0.3]{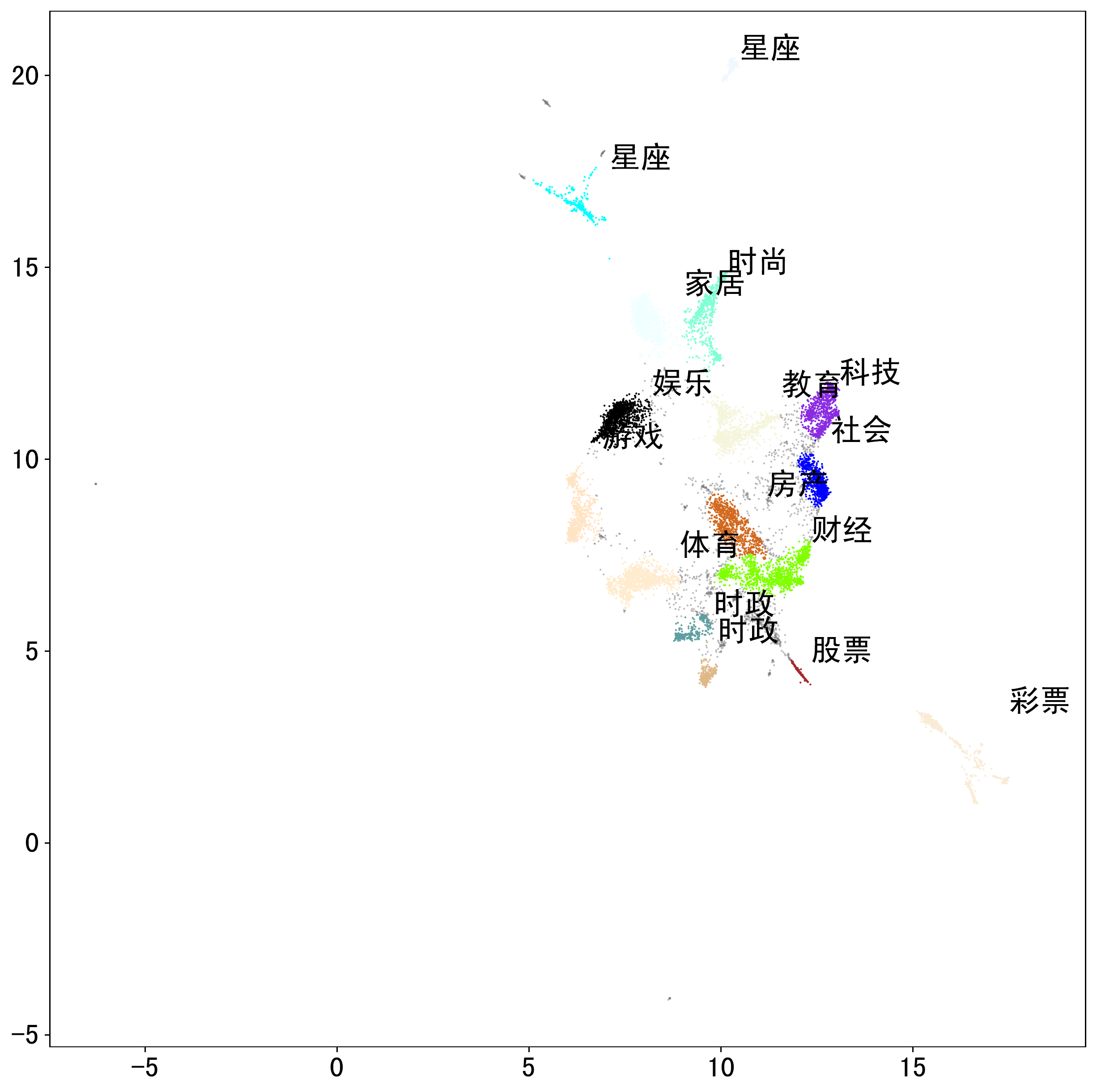}}
	\hspace{0.1in} 
	\subfigure[BERT-enhanced+UMAP+HDBSCAN]{
		\includegraphics[scale=0.3]{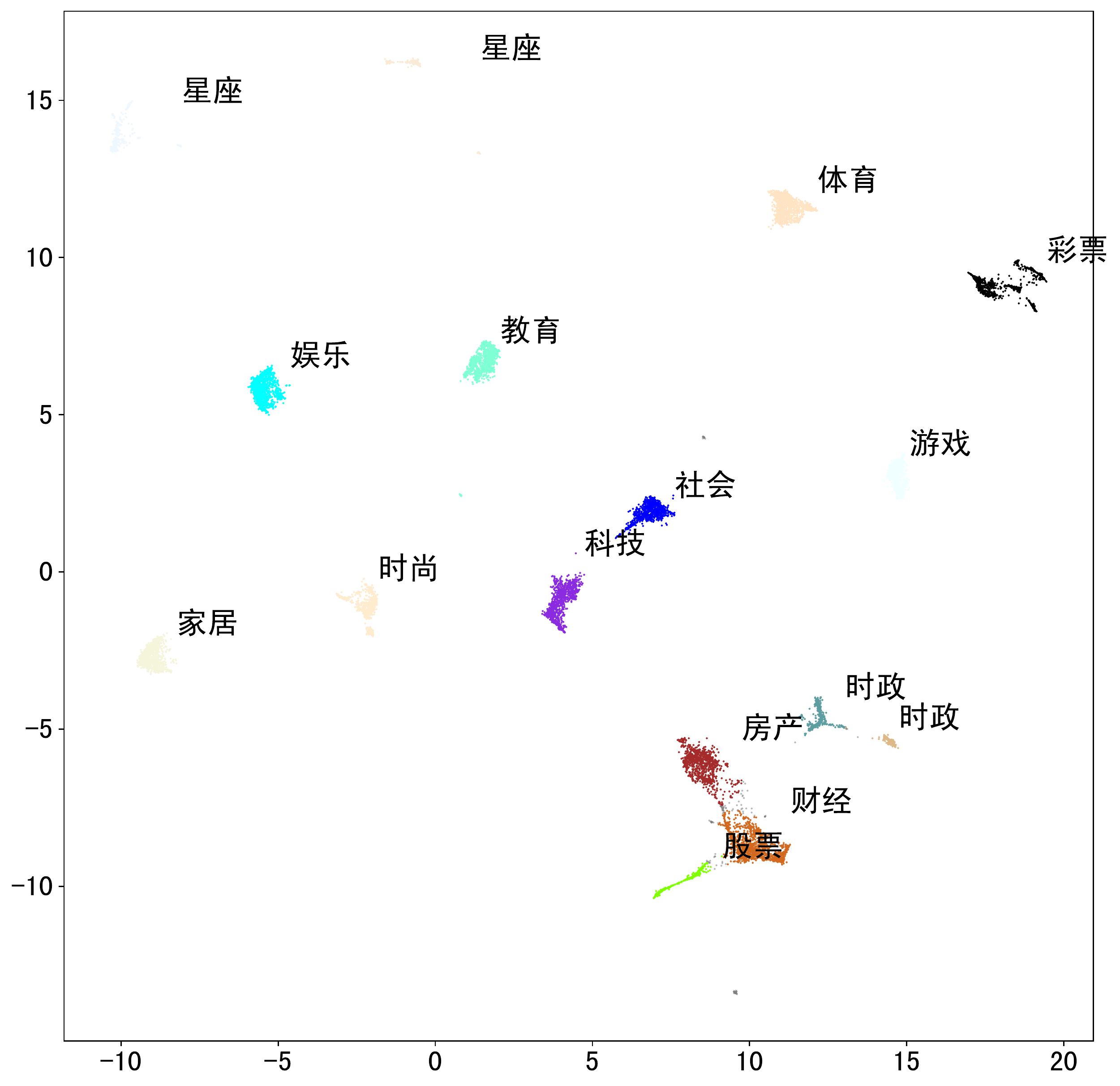}}
	
	\subfigure[SimCSE+UMAP+HDBSCAN]{
		\includegraphics[scale=0.3]{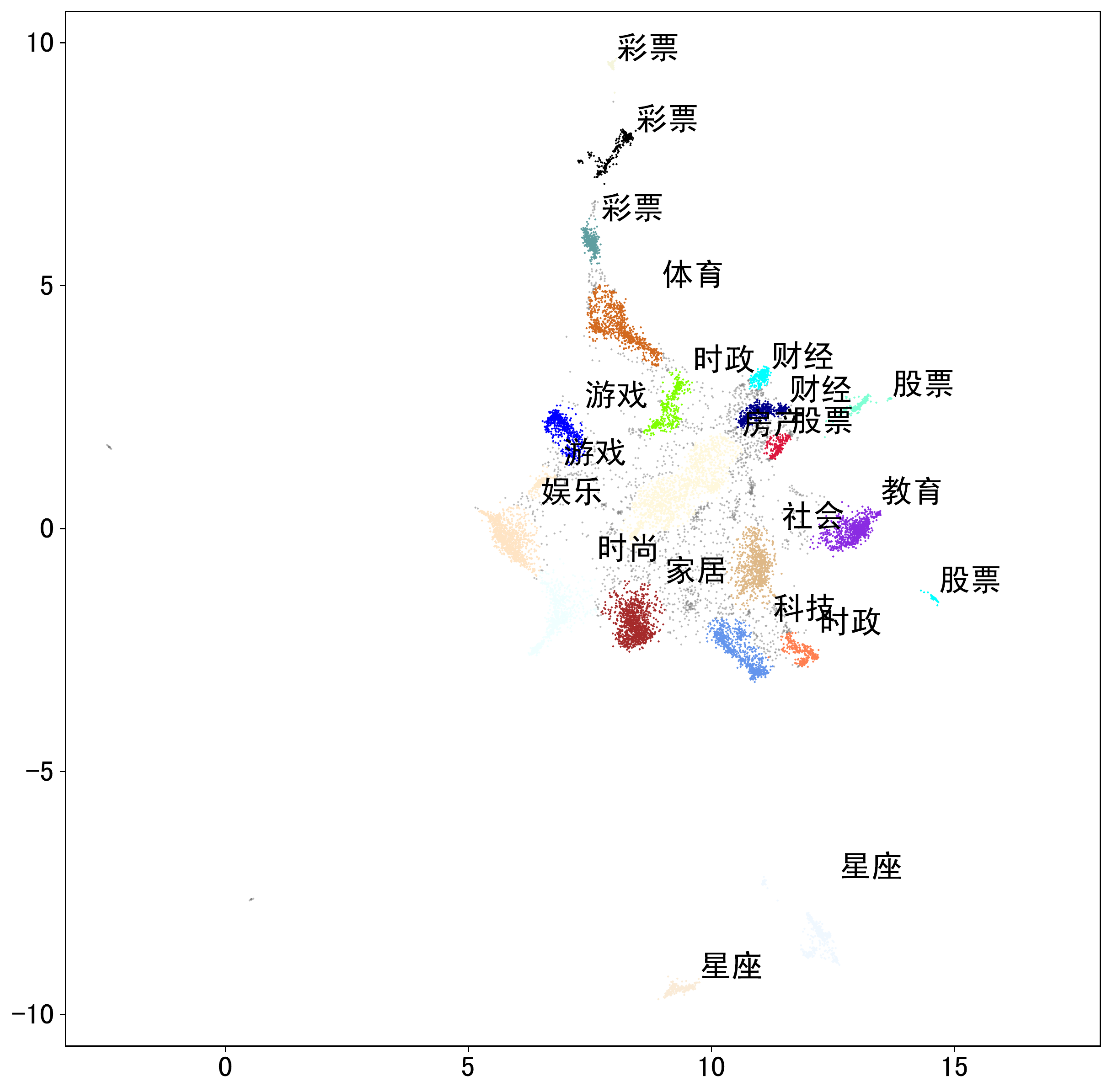}}
	\hspace{0.1in} 
	\subfigure[SimCSE-enhanced+UMAP+HDBSCAN]{
		\includegraphics[scale=0.3]{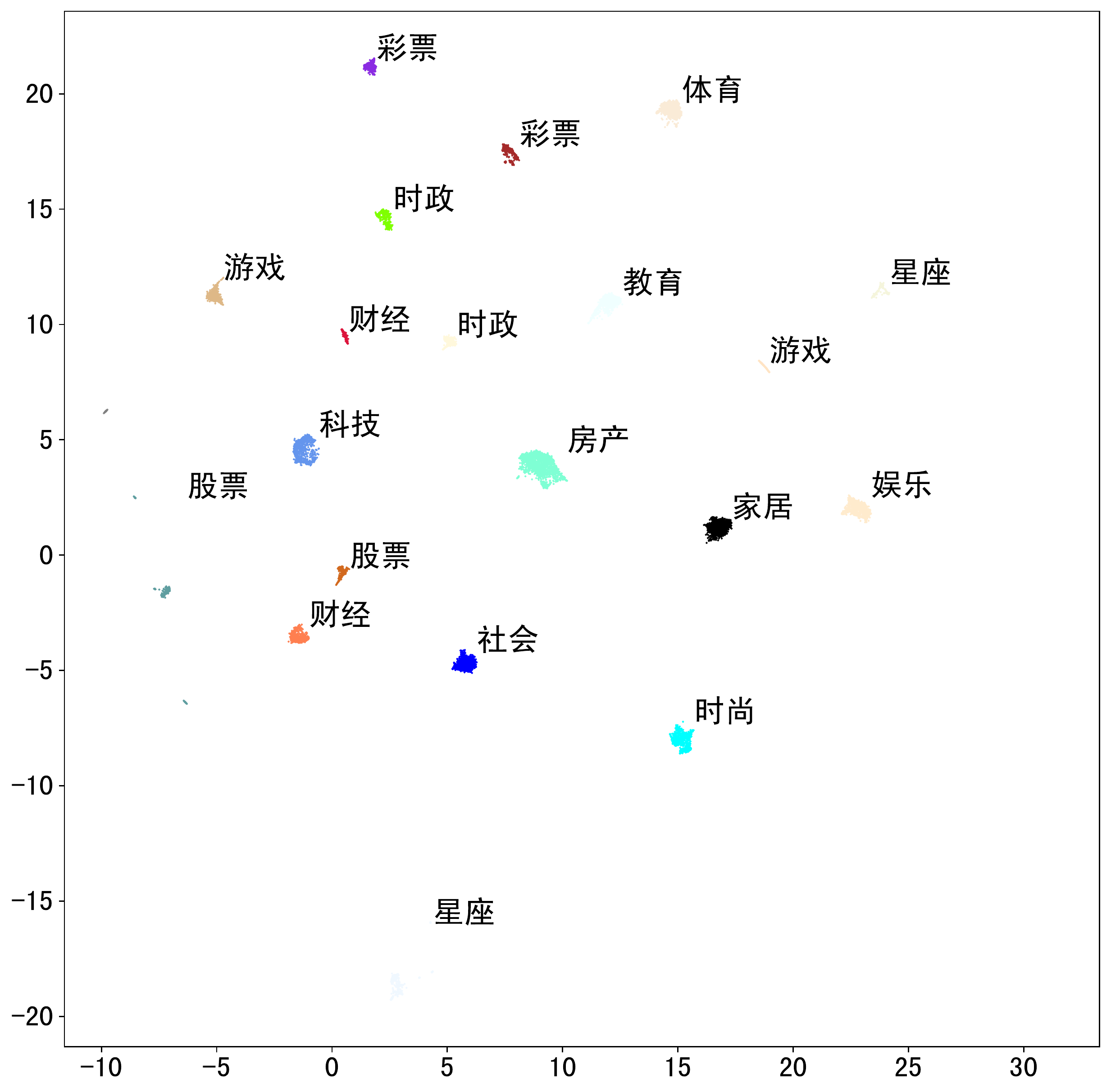}}
	\captionsetup{font=scriptsize}\caption{The comparison of clustering effects before and after the enhancement for BERT and SimCSE on THUCNews dataset. The following are a translations of the Chinese words in the picture: 彩票(lottery), 体育(sports), 时政(political science), 财经(finance), 游戏(game), 股票(stock), 房产(real estate), 娱乐(entertainment), 社会(society), 教育(education), 时尚(fashion), 家居(home furnishings), 科技(technology), 星座(constellation).}
\label{Figure-00}
\end{figure}

Figure \ref{Figure-00} shows the clustering results of the text representation obtained before and after BERT(or SimCSE) is enhanced when the UMAP and HDBSCAN combined algorithms are fixed.
It can be seen that compared with the original language model, the boundaries of the clusters in the embedding space obtained by the enhanced language model are more obvious, the distance between different clusters is farther, and the points in the same cluster are more compact.
In addition, the clusters obtained by the enhanced SimCSE are more uniform in space than those obtained by the enhanced BERT, and the shape is more similar to a sphere, which further verifies the conclusion in Table \ref{Table-1} that the internal evaluation metrics of SimCSE are better than BERT.\par

Since the THUCNews ground truth can only reflect one possible division of the dataset, which doesn't exclude the possibility of better divisions, and considering the influence of different combinations of clustering numbers, the fact that BERT outperforms SimCSE in all external evaluation metrics does not necessarily mean that its actual clustering ability is better.
In real-world scenarios, since the data generally has no real labels, when we divide the data into categories, we consider more the distribution characteristics of the data itself, and we hope to get the embedding results like the enhanced SimCSE in Figure 3, which is evenly distributed in each category. 
Therefore, for the data in the real scenarios, we chose the SimCSE with better internal evaluation metrics for text embedding.
Table \ref{Table-2} shows the clustering results of SimCSE and enhanced SimCSE on the social governance dataset. 
It can be seen that the combination that contains the enhanced language model greatly improves the recall rate and F1 value. 
This is mainly due to the fact that the enhanced language model can improve the cluster representation ability of text semantics, reduce the number of edge points between clusters in the original representation space, and reduce the generation of noise points during the HDBSCAN clustering process. 
The social governance dataset is annotated by manually judging the rationality of the clustering results.

\begin{table}[htbp]
	\centering
	\begin{threeparttable}
		\resizebox{\textwidth}{!}{
			\begin{tabular}{lccccccc}
				\hline
				\textbf{model}             & \textbf{ARI}                  & \textbf{AMI}         & \textbf{Purity}      & \textbf{F1}		& \textbf{SC}		& \textbf{CHI}		& \textbf{DBI}\\ \hline
				LDA                        & 0.021712650                    & 0.043507083          & 0.156500000                & 0.094070374     & \textbf{-}		& \textbf{-}		& \textbf{-}     \\
				FastText                   & 0.036345435                    & 0.279554411          & 0.476142857          & 0.116871716		& 0.165747600		& 263.7635306		& 1.325474649          \\ 
				Top2Vec                    & 0.053700707                    & 0.189647257          & 0.306285714          & 0.139114747		& 0.059875946      & 1358.713134		& 1.078640084    \\ \hline
				BERT\tnote{1}+PCA+KMeans         & 0.112950249                    & 0.263254817          & 0.303714286          & 0.179526423    & 0.333537489		& 10210.14964		& 0.811310507\\
				BERT+PCA+DBSCAN            & 0.030189282                    & 0.126263402          & 0.174285714          & 0.147755171		& -0.514981608		& 41.01350480		& 1.459738375          \\
				BERT+PCA+HDBSCAN           & 0.003900224                    & 0.116886870           & 0.188928571          & 0.123785626		& -0.447570568		& 36.22244089		& 1.502777369          \\
				BERT+t-SNE+KMeans        & \textbf{0.658144084}         & \textbf{0.732398618} & \textbf{0.807000000}         & \textbf{0.682909235}   & 0.484918600		& 19709.90657		& 0.698602062\\
				BERT+t-SNE+DBSCAN          & 0.562179536                   & 0.711425242          & 0.721857143          & 0.595534269		& 0.312350720		& 6251.544325		& 1.871509697 \\
				BERT+t-SNE+HDBSCAN         & 0.522267486                     & 0.697662961          & 0.665357143          & 0.560068498		& 0.298417870		& 5731.753357		& 1.553215702         \\
				BERT+UMAP+KMeans          & 0.594205812                     & 0.723575557          & 0.706785714          & 0.627605676          & \textbf{0.589915900}		& \textbf{31211.98794}		& \textbf{0.472229917}\\
				BERT+UMAP+DBSCAN           & 0.571411301                    & 0.711886810          & 0.780357142          & 0.602886097		& 0.5070201000		& 2681.931503		& 1.744340103          \\
				BERT+UMAP+HDBSCAN\tnote{2}          & \textbf{0.642479532}  & \textbf{0.740278657} & \textbf{0.837714286} & \textbf{0.667820329}		& 0.510587200		& 4447.298455		& 1.722989958  \\
				SimCSE\tnote{2}+PCA+KMeans        & 0.106347177                    & 0.252775170           & 0.288357143          & 0.175183474    & 0.344200357		& 11244.445811		& 0.815459048\\
				SimCSE+PCA+DBSCAN          & 0.044355821                    & 0.151427560           & 0.159000000                & 0.164180209		& -0.254810313		& 230.1634931		& 2.394114761           \\
				SimCSE+PCA+HDBSCAN         & 0.002175732                    & 0.113755726          & 0.192142857          & 0.123710760		& -0.424716346		& 35.80529592		& 1.956265486           \\
				SimCSE+t-SNE+KMeans       & 0.547095723                    & 0.661444030           & 0.729714286          & 0.581130327		& 0.474737140		& 18324.75306		& 0.645366147\\
				SimCSE+t-SNE+DBSCAN        & 0.531404366                   & 0.679330645          & 0.784571429          & 0.562565842		& 0.360822170		& 5695.250806		& 1.470280551            \\
				SimCSE+t-SNE+HDBSCAN       & 0.380231096                   & 0.653010575          & 0.726071429          & 0.429845375		& 0.282251420		& 3646.129072		& 1.513032985          \\
				SimCSE+UMAP+KMeans        & 0.498917295                    & 0.650605537          & 0.659714286          & 0.540381135		& \textbf{0.540319100}		& \textbf{24314.62983}		& \textbf{0.491694285}            \\
				SimCSE+UMAP+DBSCAN         & 0.552333818                    & 0.679338022   & 0.792357143  & 0.582409120		& 0.521376670		& 5455.263584		& 1.182661660              \\
				SimCSE+UMAP+HDBSCAN        & 0.509769071                    & 0.674170452          & 0.790928571          & 0.542680251		& 0.480586700		& 5832.627975		& 1.805907217            \\ \hline
				BERT-enhanced+PCA+KMeans     & 0.446185093                    & 0.577556915          & 0.597642857          & 0.489695836          & 0.513355760		& 28930.99599		& 0.703627083\\
				BERT-enhanced+PCA+DBSCAN      & 0.306394174                   & 0.528642067          & 0.452000000                & 0.374108656		& 0.388692892		& 6922.794374		& 1.276219364          \\
				BERT-enhanced+PCA+HDBSCAN     & 0.270715954                    & 0.530203903          & 0.528785714          & 0.340734720		& 0.182775500		& 3178.215554		& 1.462026462          \\
				BERT-enhanced+t-SNE+KMeans  & 0.669056669                    & 0.744554080           & 0.816142857          & 0.693527327           & 0.634305800		& 35938.15804		& 0.500305803\\
				BERT-enhanced+t-SNE+DBSCAN    & \textbf{0.713460719}          & 0.768681769          & 0.872357142      & 0.732967082		& 0.623143850		& 21405.49619		& 1.457663276 \\
				BERT-enhanced+t-SNE+HDBSCAN   & \textbf{0.719518003}            & 0.771625801          & \textbf{0.875285714}          & \textbf{0.738656874}		& 0.631086800		& 26618.08005		& 0.783802092        \\
				BERT-enhanced+UMAP+KMeans    & 0.698639960            & \textbf{0.774216205} & 0.830785714          & 0.721363204		& 0.808400150		& 157304.9432		& \textbf{0.247107253}                   \\
				BERT-enhanced+UMAP+DBSCAN     & 0.711797510                     & 0.771620123          & 0.870642857          & 0.731856021		& 0.821047100		& 26382.62055		& 2.287925582         \\
				BERT-enhanced+UMAP+HDBSCAN    & 0.718627893                     & \textbf{0.773906464}   & \textbf{0.877428571} & \textbf{0.738152439}		& 0.830618300		& 98233.07291		& 1.142740436 \\
				SimCSE-enhanced+PCA+KMeans   & 0.391048446                    & 0.501399593          & 0.564714286          & 0.438903832		& 0.550941828		& 45130.16688		& 0.659255520              \\
				SimCSE-enhanced+PCA+DBSCAN    & 0.230694030                     & 0.500170841          & 0.500571429          & 0.305328603		& 0.252246800		& 3402.093261		& 1.321582415              \\
				SimCSE-enhanced+PCA+HDBSCAN   & 0.290785881                    & 0.512756342          & 0.536500000               & 0.356713658		& 0.362360205		& 11606.72156		& 2.352973108          \\
				SimCSE-enhanced+t-SNE+KMeans & 0.524010214                    & 0.649913615          & 0.703285714          & 0.559648674		& 0.672626400		& 31379.05478		& 0.542381072           \\
				SimCSE-enhanced+t-SNE+DBSCAN  & 0.586704695                   & 0.705531691          & 0.842071429          & 0.614094205		& 0.785797830		& 83140.03355		& 1.260080529           \\
				SimCSE-enhanced+t-SNE+HDBSCAN & 0.590634454                    & 0.704936408          & 0.843714286          & 0.617538117		& 0.760994260		& 84049.29390		& 0.401127394         \\
				SimCSE-enhanced+UMAP+KMeans  & 0.584538557                    & 0.687672859          & 0.755071429          & 0.615344342		& 0.788089900		& 42994.13445		& 0.455941761           \\
				SimCSE-enhanced+UMAP+DBSCAN   & 0.587115673                    & 0.705317471          & 0.842214286          & 0.614454433		& \textbf{0.919167700}		& \textbf{285574.0943}		& 0.378244737          \\
				SimCSE-enhanced+UMAP+HDBSCAN  & 0.591524314                    & 0.707746769          & 0.842214286          & 0.618695607		& \textbf{0.912691300}		& \textbf{269272.5169}		& \textbf{0.163748030}          \\ \hline
			\end{tabular}
		}
	\end{threeparttable} 
	\captionsetup{font=scriptsize}\caption{ Evaluation of clustering on THUCNews dataset. It shows the performance of three benchmark models and the combination of different models on the THUCNews dataset under the framework of text representation, dimensionality reduction, and clustering. The parameters of all dimensionality reduction and clustering algorithms adopt the optimal parameters according to the above parameter selection methods. In all combinations, KMeans selects the true category number 14 of the THUCNews dataset as the k value. "BERT" represents the BERT-base-chinese model, "SimCSE" represents the SimCSE-chinese-roBERTa model, "enhanced" represents the enhanced model, and "BERT+UMAP+HDBSCAN" is the combination method used by BERTopic. In the table, the data of the top two metrics before and after the improvement are respectively bolded. Among the evaluation metrics, only the smaller the DBI is, the better the clustering result is, and vice versa for other metrics.}
	\label{Table-1}
\end{table}

\begin{table}[htbp]\footnotesize
	\centering
	\resizebox{\textwidth}{!}{
	\begin{tabular}{lcccc}
    \hline
    \textbf{model}            & \textbf{Precision} & \textbf{Recall} & \textbf{Accuracy} & \textbf{F1} \\ \hline
    SimCSE+UMAP+HDBSCAN       & \textbf{0.938622883}        & 0.68359106      & 0.647243628       & 0.791059915 \\ 
    SimCSE-enhanced+UMAP+HDBSCAN & 0.923798212        & \textbf{0.946822142}     & \textbf{0.875992887}       & \textbf{0.935168486} \\ \hline
   \end{tabular}
   }	
	\captionsetup{font=scriptsize}\caption{The clustering results on social governance dataset. The table shows the results before and after the enhancement of the pre-trained language model. SimCSE represents the SimCSE-chinese-roBERTa model, and "enhanced" represents the enhanced model. The maximum value of each metric is bolded.}
\label{Table-2}
\end{table}

\subsection{Topic Extraction}

Table \ref{Table-4} shows the topics learned by the proposed attention based method and the c-TF-IDF algorithm in BERTopic on the social governance dataset and the THUCNews dataset, where the attention mechanism comes from BERT, SimCSE and their enhanced versions . In order to emphasize the comparison of different topic extraction methods, this paper selects 5 types of labeled data from the social governance dataset and the THUCNews dataset respectively forms two sub-datasets with real labels. The enhanced BERT and SimCSE in Table \ref{Table-4} obtained by training the classification model for 20 epochs on the sub-datasets. It can be seen from Table \ref{Table-4} that compared with c-TF-IDF, the top 5 keywords extracted by the attention mechanism method based on the enhanced language model can better reflect the central semantics of the category. For example, c-TF-IDF extracted a large number of meaningless words on the social governance dataset, such as "小区(residential complex)", "居民(resident)", "社区(community)" and "情况(situation)", etc., while the words extracted by SimCSE-enhanced+Attention were all related to the topic. Although BERT-enhanced+Attention also extracted some words that were not related to the category topics, such as "影响(influence)" and "双方(both sides)", its overall effect is still better than c-TF-IDF. 

Although c-TF-IDF adopts class-based TF-IDF, which enables it to reduce the importance ranking of high-frequency words that appear in multiple categories at the same time, it is limited in the following two aspects. 
First, when the number of categories is small, the role of idf will be weakened. 
For example, "组图(pictures)" appears in three categories "电影(movie)", "留学(study abroad)" and "科学(science)" at the same time. However, the total number of categories is 5, which is relatively small, so the influence of idf is small, and the influence of high word frequency cannot be avoided in the end. 
Secondly, if there is a category-independent high-frequency word in a category and it does not appear in other categories, the high-frequency word will still have a great impact on the result, and can be easily mistaken for the central word of the category. 
For example, "纠纷(dispute)" is a high-frequency word in the category of "土地(land)", but it does not appear in a large number of other categories, so its weight of idf is very small.
Finally, it will be mainly affected by the word frequency of "纠纷(dispute)", and it will be considered as the keyword of the category. 
The advantage of the attention based topic extraction is that in the process of getting the enhanced language model (the training of classification model), the model has learned the distinctions between categories, so that the enhanced language model shifts its attention to the words related to the categories. 
For example, the language model before enhancement is more likely to pay attention to "纠纷(dispute)", but because there are "纠纷(dispute)" in different categories, it means that "纠纷(dispute)" is the common feature of different categories. 
In the training process of the enhanced language model, its attention to the common feature will gradually weaken, and finally transfer its attention to the representative keywords that can distinguish different categories. 
It avoids the phenomenon of extracting irrelevant words such as "纠纷(dispute)", "居民(resident)" and "组图(pictures)".\par

In the four combinations based on the attention mechanism, the enhanced language model version extracted a larger proportion of topic-related words than the original version.
As can be seen from Table \ref{Table-4}, BERT (or SimCSE) + Attention extracted 9 (or 4) category-irrelevant words on the five categories of the social governance dataset, while BERT (or SimCSE) -enhanced + Attention only extracted 2 (or 0) words. 
As the language model is enhanced, the importance ranking of keywords has also changed. 
For example, after SimCSE is enhanced, its focus on the category "丢失物(lost property)" has changed from the word "钥匙(key)" to the word "丢失(lost)" that is most related to the category topic. 
Besides, Table 4 also shows that the language model SimCSE outperforms BERT on average. 

Since both of the proposed attention based topic extraction and c-TF-IDF methods rely on sentence tokenization, affected by the tokenizer, the extracted keywords might contain nonstandard words such as " 绳狗屎" and "冠疫苗".

\begin{table}[htbp]
	\centering
	\resizebox{\textwidth}{!}{
\begin{tabular}{l|ccccc|ccccc}
	\hline
	\multirow{3}{*}{\textbf{Method}}         
	& \multicolumn{5}{c|}{\textbf{Social governance dataset}}  & \multicolumn{5}{c}{\textbf{THUCNews}}  
	\\ \cline{2-11} 
	&\textbf{遛狗}   &\textbf{丢失物}  & \textbf{疫苗}  & \textbf{噪音}  & \textbf{土地} 
	& \textbf{足球}  &\textbf{电影}    & \textbf{彩票}  & \textbf{留学}  & \textbf{科学}
	\\
	&\textbf{(walking the dog)}   &\textbf{(lost property)}  & \textbf{(vaccine)}  & \textbf{(noise)}  & \textbf{(land)}
	& \textbf{(soccer)}  &\textbf{(movie)}    & \textbf{(lottery)}  & \textbf{(study abroad)}  & \textbf{(science)}
	\\ \hline
	\multirow{10}{*}{groudtruth classes + c-TF-IDF}           
	& 遛狗  & 钥匙    & 接种  & 休息  & \underline{纠纷} 
	& 足协  & 电影    & 负彩  & 留学  & 科学  
	\\
	& (walking the dog)  & (key)    & (inoculate)  & (rest)  & \underline{(dispute)} 
	& (Football Association)  & (movie)    & (negative lottery)  & (study abroad)  & (science)  
	\\
	& \underline{小区} & 丢失 & 疫苗 & \underline{影响} & \underline{发生} 
	& 女足             & 票房 &\underline{推荐} & 美国  & 发现
	 \\
	& \underline{(residential complex)} & (lose) & (vaccine) & \underline{(affect)} & \underline{(happen)} 
	& (women's football)             & (box office) &\underline{(recommend)} & (America)  & (discovery) 
	\\
	& 狗绳 & \underline{社区} & 冠疫苗 & 楼上 & \underline{产生}
	& 中国 & \underline{香港} & 主场 & \underline{组图} & \underline{组图} 
	\\
	& (dog leash) & \underline{(community)} & (COVID-19 vaccine) & (upstairs) & \underline{(produce)}
	& (China) & \underline{(Hong Kong)} & (home court) & \underline{(pictures)} & \underline{(pictures)} 
	\\
	& \underline{居民} & 身份证 & \underline{咨询} & \underline{居民} & 土地       
	& 足球             & 新片   & 开奖             & 大学             & 地球
	\\
	& \underline{(resident)} & (identity card) & \underline{(inquire)} & \underline{(resident)} & (land)       
	& (soccer)             & (new movie)   & (draw a lottery)             & (university)             & (earth)         
	\\
	& 栓绳 & 失主  & \underline{情况} & 扰民       & \underline{问题} 
	& 韦迪 & \underline{组图} & 火线 & 移民       & \underline{英国} 
	\\
	& (tether) & (owner of lost property)  & \underline{(situation)} & (nuisance)       & \underline{problem} 
	& (Di Wei) & \underline{(pictures)} & (firing line) & (emigrate)       & \underline{(Britain)} 
	\\ \hline
	\multirow{10}{*}{BERT+Attention}         
	& 拴绳       & 丢失             & 冠疫苗      & 扰民             & \underline{纠纷} 
	& 足协       & \underline{组图} & 双色球      & \underline{组图} & \underline{组图} 
	\\
	& (tether)       & (lose)             & (COVID-19 vaccine)      & (nuisance)             & \underline{(dispute)} 
	& (Football Association)       & \underline{(pictures)} & (double chromosphere)      & \underline{(pictures)} & \underline{(pictures)} 
	\\
	& 遛狗       & 失主 & 疫苗       & \underline{影响} & 地界       
	& 女足       & 票房  & 彩民       & 留学 & 月球 
	\\
	& (walking the dog)       & (owner of lost property) & (vaccine)       & \underline{(affect)} & (boundary)       
	& (women's football)       & (box office)  & (lottery buyer)       & (study abroad) & (moon)        
	\\
	& 狗绳 & 身份证  & \underline{咨询} & 吵闹 & \underline{村委会}      
	& 足球 & 电影 & 头奖 & 移民    & \underline{2008}  
	\\
	& (dog leash) & (identity card)  & \underline{(inquire)} & (wrangle) & \underline{(village committee)}      
	& (soccer) & (movie) & (the first prize) & (emigrate)    & \underline{(2008)}     
	\\
	& 绳子       & 钥匙       & \underline{情况} & 噪音 & \underline{村民}      
	& 球员       & 甄子丹       & 初盘             & 澳洲  & 揭秘       
	\\
	& (rope)       & (key)       & \underline{(situation)} & (noise) & \underline{(villager)}      
	& (players)       & (Donnie Yen)       & (initial game)             & (Australia)  & (reveal) 
	\\
	& 绳狗屎       & 忘记 & \underline{询问} & \underline{孩子} & \underline{争执} 
	& 足坛       & \underline{打分} & 负彩 & 美国       & 火星   
	\\
	& (rope dog poop) & (forget) & \underline{(inquire)} & \underline{(child)} & \underline{(dispute)} 
	& (football world)       & \underline{(score)} & (negative lottery) & (America)       & (Mars)       
	\\ \hline
	\multirow{10}{*}{BERT-enhanced+Attention}   
	& 遛狗       & 钥匙 & 疫苗       & \underline{影响} & 土地 
	& 女足       & 梅兰芳       & \underline{推荐} & 留学 & 科学      
	\\
	& (walking the dog)       & (key) & (vaccine)       & \underline{(affect)} & (land) 
	& (women’s football)       & (Forever Enthralled)       & \underline{(recommend)} & (study abroad) & (science)      
	\\
	& 狗绳       & 丢失       & 接种      & 噪音       & 宅基地
	& 足协       & 画皮       & 火线      & 美国      & 发现 
	\\
	& (dog leash)       & (lose)       & (inoculate)      & (noise)       & (homestead)
	& (Football Association)       & (Painted.Skin)       & (firing line)      & (America)      & (discovery) 
	\\
	& 狗链       & 身份证       & 冠疫苗      & 声音       & \underline{双方}      
	& 足球       & 冯小刚       & 双色球      & 留学生      & 火星     
	\\
	& (dog leash)       & (identity card)       & (COVID-19 vaccine)      & (voice)       & \underline{(both sides)}      
	& (soccer)       & (Xiaogang Feng)       & (double chromosphere)      & (overseas student)      & (Mars)     
	\\
	& 绳子 & 失主 & 打疫苗 & 吵闹 & 地界       
	& 世界杯  & 叶问    & 负彩 & 英国 & 研究   
	\\
	& (rope) & (owner of lost property) & (get vaccinated) & (wrangle) & (boundary)       
	& (world cup)  & (IP MAN)    & (negative lottery) & (Britain) & (study)     
	\\
	& 绳狗屎 & 钱包  & 针疫苗   & 晚上 & 占地      
	& 中国 & 海角七号 & 彩票  & 出国 & 太空
	\\
	& (rope dog poop) & (wallet)  & (nth-dose vaccine)  & (night) & (appropriation of land)      
	& (China) & (Cape No.7) & (lottery)  & (go abroad) & (space)        
	\\ \hline
	\multirow{10}{*}{SimCSE+Attention}       
	& 遛狗       & 钥匙       & 疫苗       & 扰民       & \underline{纠纷} 
	& 女足       & 票房       & 双色球      & 留学      & 火星       
	\\
	& (walking the dog)       & (key)       & (vaccine)       & (nuisance)       & \underline{(dispute)} 
	& (women’s football)       & (box office)       & (double chromosphere)      & (study abroad)      & (Mars)       
	\\
	& 拴绳       & 丢失       & 冠疫苗      & 噪音       & 宅基地      
	& 足协      & 梅兰芳      & 彩票        & 移民       & 机器人 
	\\
	& (tether)       & (lose)       & (COVID-19 vaccine)      & (noise)       & (homestead)      
	& (Football Association)      & (Forever Enthralled)      & (lottery)        & (emigrate)       & (robot) 
	\\
	& 牵绳       & 失主       & \underline{咨询}      & \underline{影响}   & 占地      
	& 足球       & 赤壁       & \underline{米兰}      & 签证      & 黑洞      
	\\
	& (leash)   & (owner of lost property)       & \underline{inquire}      & \underline{(affect)}   & (appropriation of land)      
	& (soccer)       & (The War RedCliff)       & \underline{(Inter Milan)}      & (visa)      & (black hole)      
	\\
	& 狗绳      & 书包   & 接种 & \underline{邻居}  & 耕地       
	& 世界杯       & 刘烨       & \underline{排行榜}     & 申请       & 月球   
	\\
	& (dog leash)      & (schoolbag)   & (inoculate) & \underline{(neighbor)}  & (farmland)       
	& (world cup)       & (Ye Liu)       & \underline{(ranking list)}       & (apply)       & (moon)     
	\\
	& 绳子       & 被盗       & 打疫苗     & 吵闹 & 地界       
	& 主帅       & 谢霆锋       & \underline{(国米)}       & 名校       & 望远镜     
	\\
	& (rope)       & (stolen)       & (get vaccinated)     & (wrangle) & (boundary)       
	& (manager)       & (Nicholas Tse)       & \underline{(Inter Milan)}       & (prestigious university)       & (telescope)   
	\\ \hline
	\multirow{10}{*}{SimCSE-enhanced+Attention} 
	& 遛狗       & 丢失       & 疫苗       & 休息       & 土地       
	& 女足       & 梅兰芳      & 彩票       & 留学      & 科学 
	\\    
	& (walking the dog)       & (lose)       & (vaccine)       & (rest)       & (land)       
	& (women’s football)       & (Forever Enthralled)      & (lottery)       & (study abroad)      & (science)     
	\\
	& 狗绳       & 钥匙      & 冠疫苗      & 楼上       & 宅基地      
	& 足协       & 票房 & 负彩 & 美国 & 发现       
	\\
	& (dog leash)       & (key)      & (COVID-19 vaccine)      & (upstairs)       & (homestead)      
	& (Football Association)       & (box office) & (negative lottery) & (America) & (discovery)     
	\\
	& 拴绳       & 身份证     & 接种      & 扰民       & 地界       
	& 世界杯     & \underline{香港}     & \underline{推荐}      & 出国       & 研究   
	\\
	& (tether)       & (identity card)     & (inoculate)      & (nuisance)       & (boundary)       
	& (world cup)     & \underline{(Hong Kong)}     & \underline{(recommend)}      & (go abroad)       & (study)      
	\\
	& 绳子       & 失主       & 打疫苗     & 噪音       & 占地 
	& 韦迪       & 冯小刚       & 双色球     & 移民       & 南极
	\\
	& (rope)       & (owner of lost property)       & (get vaccinated)     & (noise)       & (appropriation of land) 
	& (Di Wei)       & (Xiaogang Feng)       & (double chromosphere)     & (emigrate)       & (south pole)        
	\\
	& 牵绳 & 遗失 & 针疫苗 & 晚上 & 耕地 
	& 足球 & 周迅 & 彩民 & 留学生 & 机器人  
	\\
	& (leash) & (lose) & (nth-dose vaccine) & (night) & (farmland) 
	& (soccer) & (Margaret) & (lottery buyers) & (overseas student) & (robot)     
	\\ \hline
\end{tabular}
}
\captionsetup{font=scriptsize}\caption{A comparison of the effects of different topic extraction algorithms on the social governance dataset and the THUCNews dataset. We have underlined words extracted by each algorithm that are unrelated to category topics.}
\label{Table-4}
\end{table}

\subsection{Effection of Model Enhancement}

\noindent
\textbf{Variations in the distribution of clusters}

\begin{figure}[htbp]
	\centering
	\resizebox{\textwidth}{!}{
	\subfigure[Pre-train]{
		\includegraphics[scale=0.25]{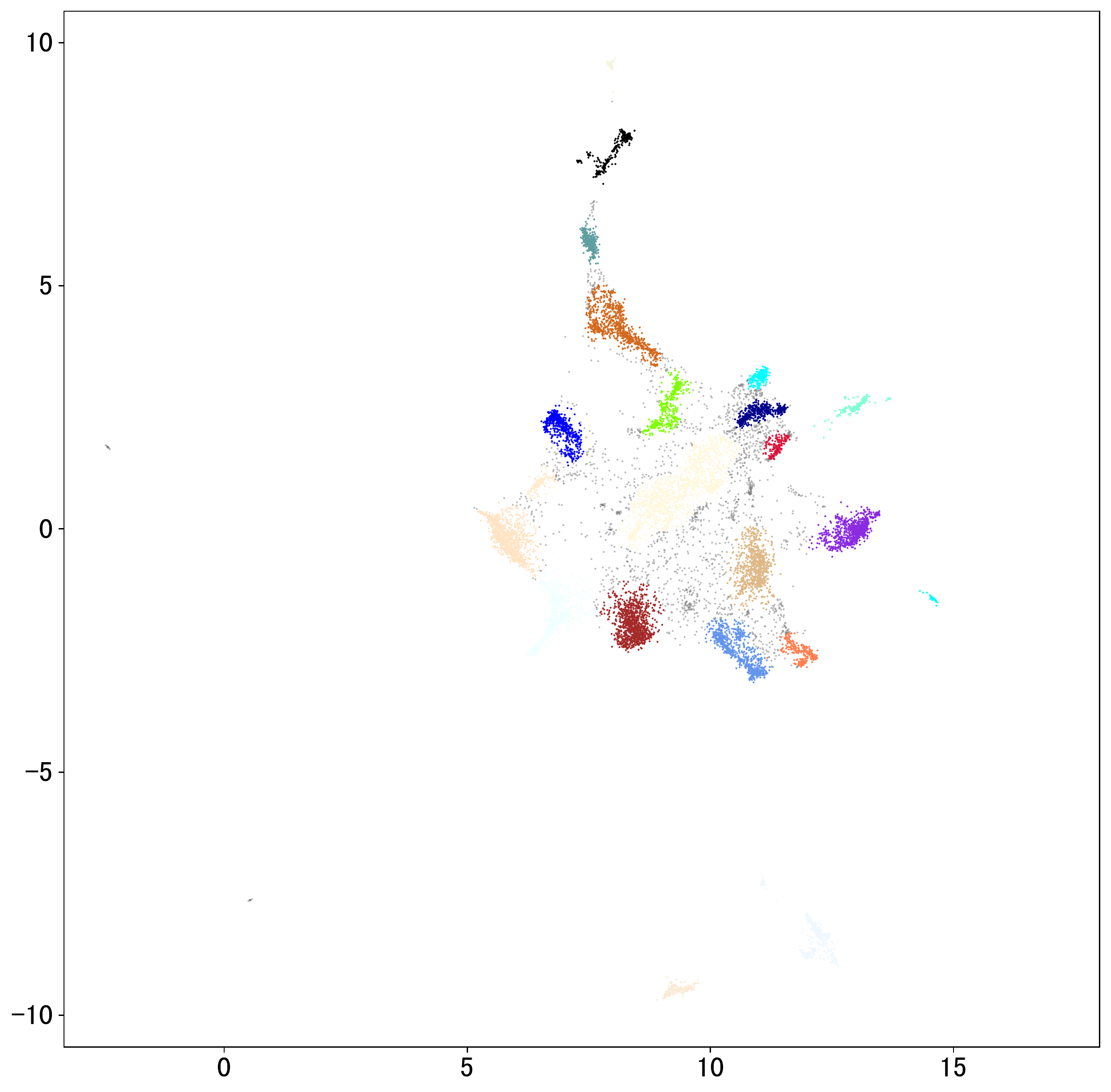}}
	\hspace{0.1in} 
	\subfigure[Epoch 0]{
		\includegraphics[scale=0.25]{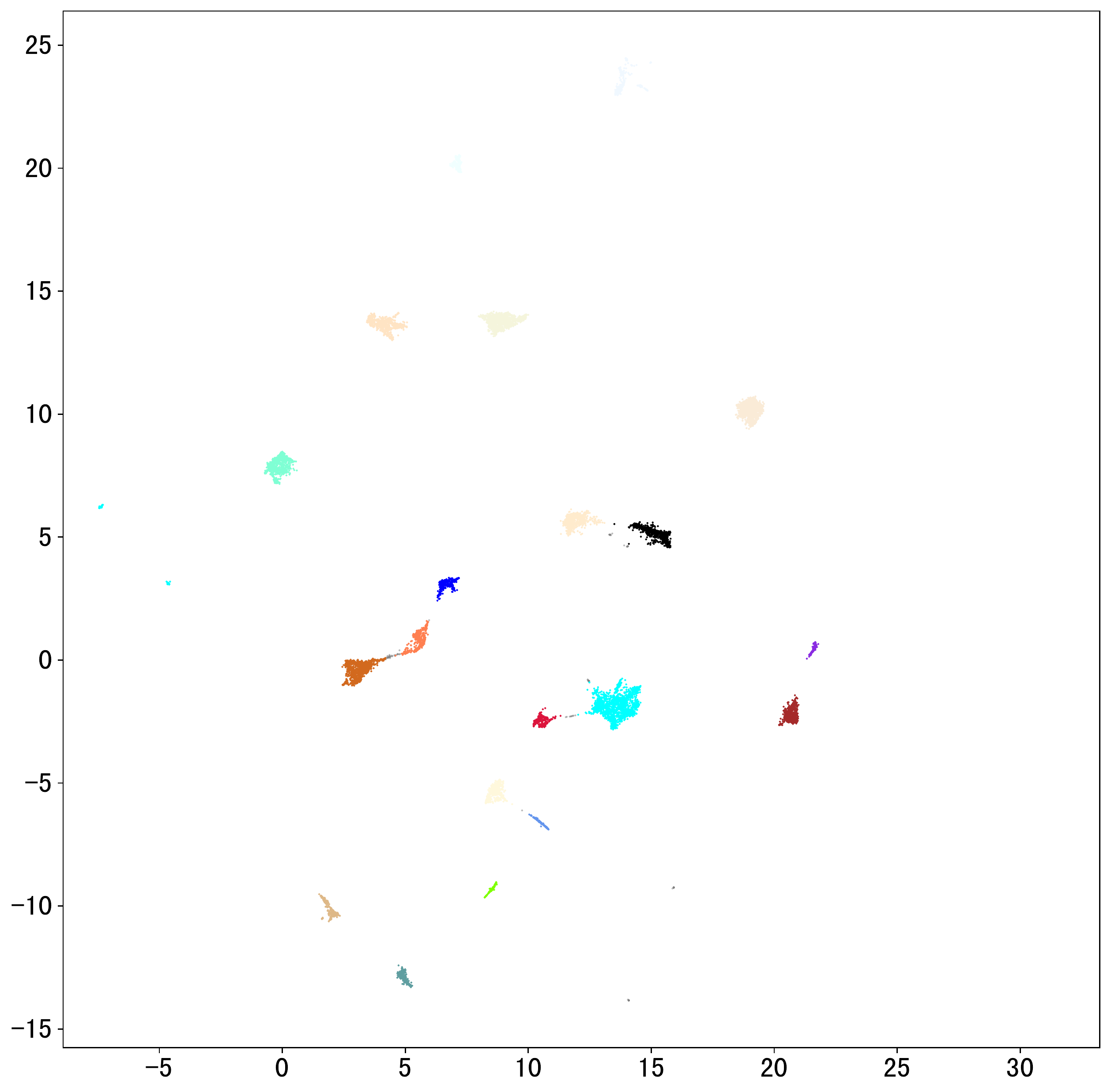}}
	}
	\resizebox{\textwidth}{!}{
	\subfigure[Epoch 1]{
		\includegraphics[scale=0.25]{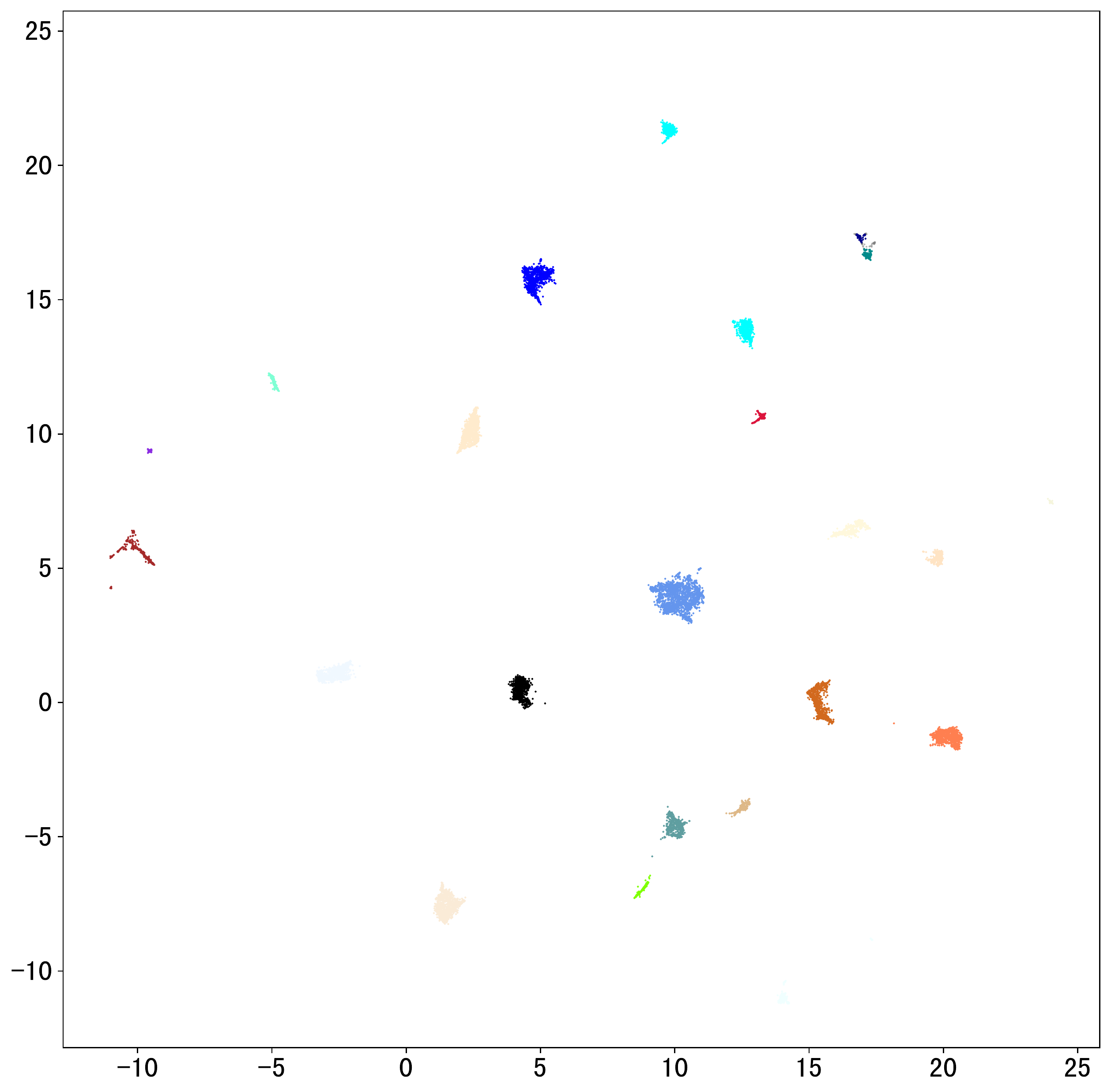}}
	\hspace{0.1in} 
	\subfigure[Epoch 3]{
		\includegraphics[scale=0.25]{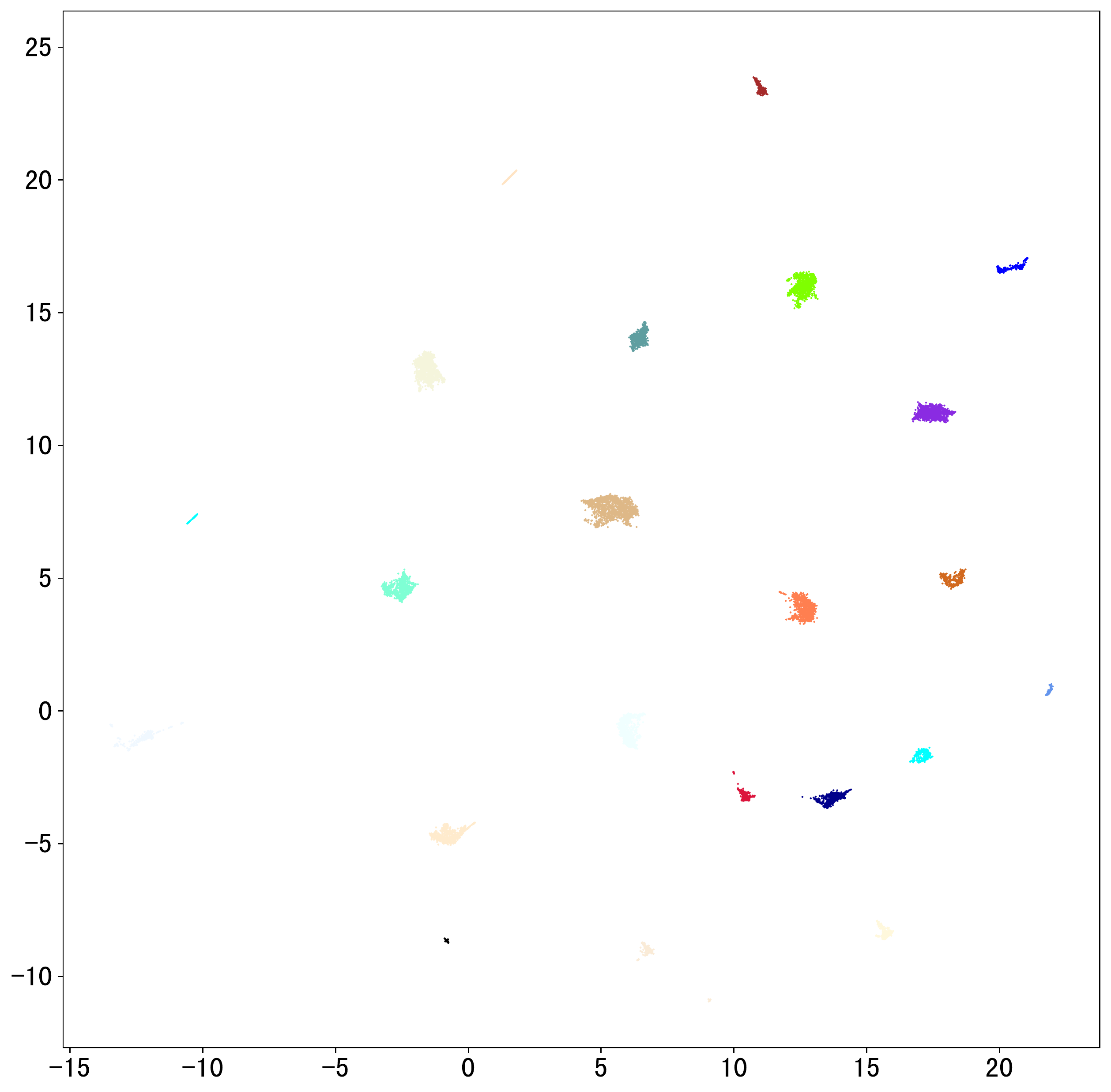}}
	}
	\captionsetup{font=scriptsize}\caption{Spatial distribution of clusters during model enhancement. The figure shows the changing trend of the spatial distribution of the 2-dimensional embeddings on the THUCNews dataset by adopting the original pre-trained SimCSE and the enhanced SimCSE. The epoch $i$($i=1,2,3$) is the model training round, and the label is the cluster label.}
\label{Figure-3}
\end{figure}

To investigate the spatial distribution variations of the embeddings during the training process, we use SimCSE to conduct experiments on the THUCNews dataset in this section, and display the variations of the embedding space in Figure \ref{Figure-3}. 

It shows that the distribution of the embeddings obtained by the original pre-trained language model does not present obvious clusters. are varied.
Moreover, there are no clear boundaries between clusters, and the cluster shapes are irregular.
With the progress of the training process, the cluster structure in the embedding space gradually becomes clearer, and the overall distribution of different clusters becomes more uniform.
Finally, a distribution with a compact intra-cluster distance and a large inter-cluster distance is generated.

\noindent
\textbf{The shift in attention}

\begin{figure}[htbp]
	\centering
	\resizebox{\textwidth}{!}{
	\subfigure[before enhancement]{
		\includegraphics[scale=0.3]{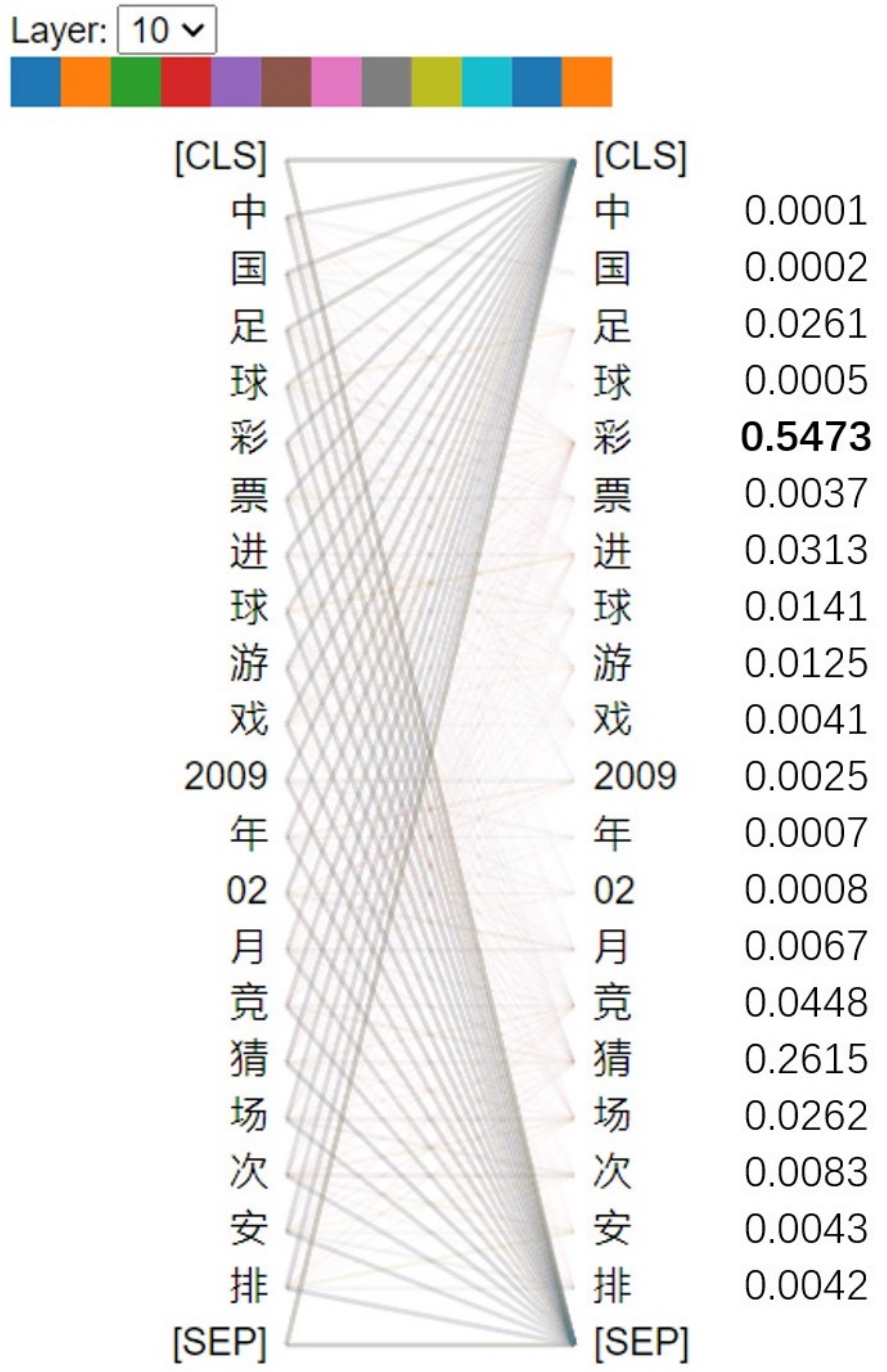}}
	\hspace{0.5in} 
	\subfigure[after enhancement]{
		\includegraphics[scale=0.3]{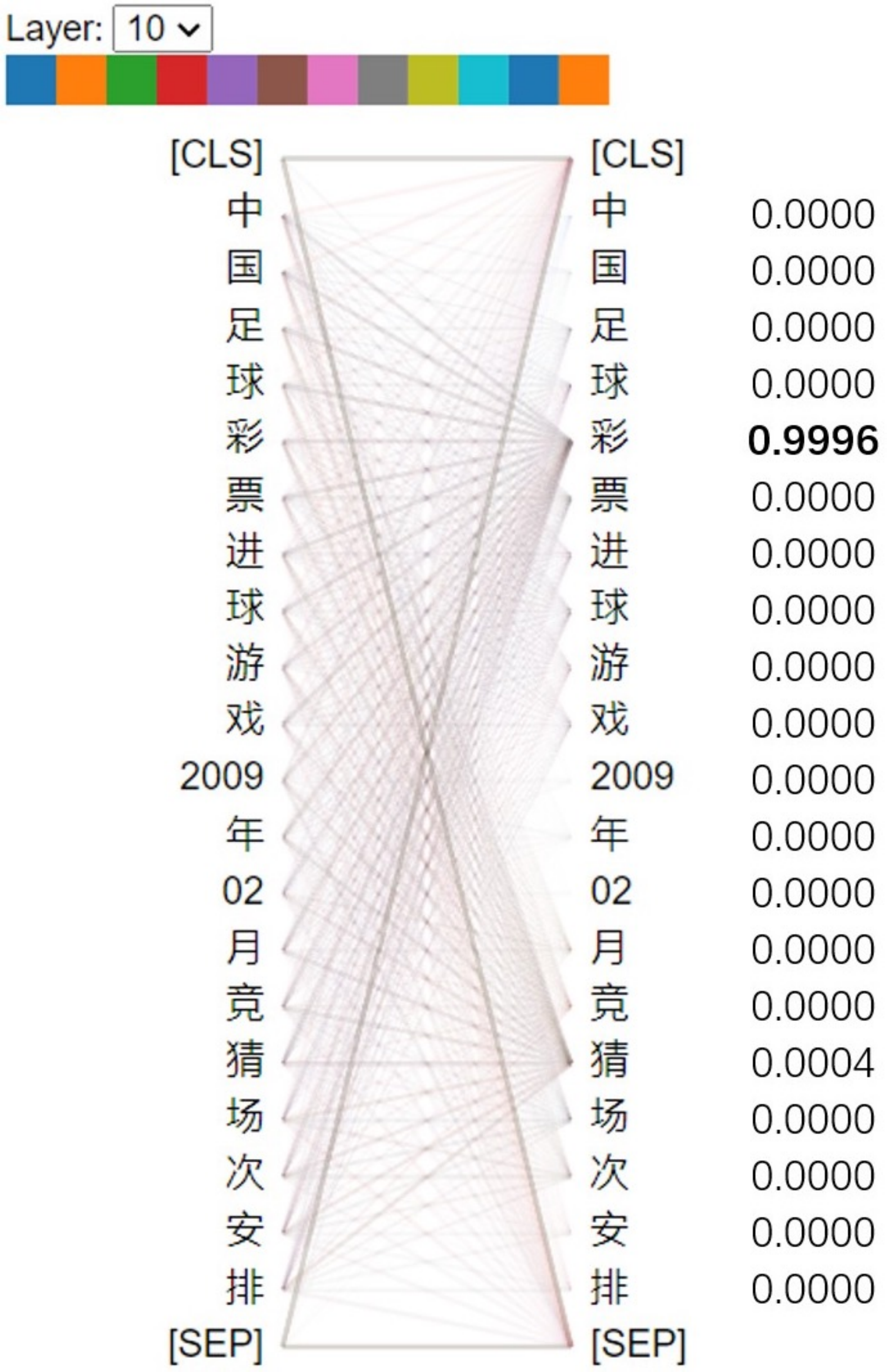}}
	}
	\resizebox{\textwidth}{!}{
	\subfigure[before enhancement]{
		\includegraphics[scale=0.3]{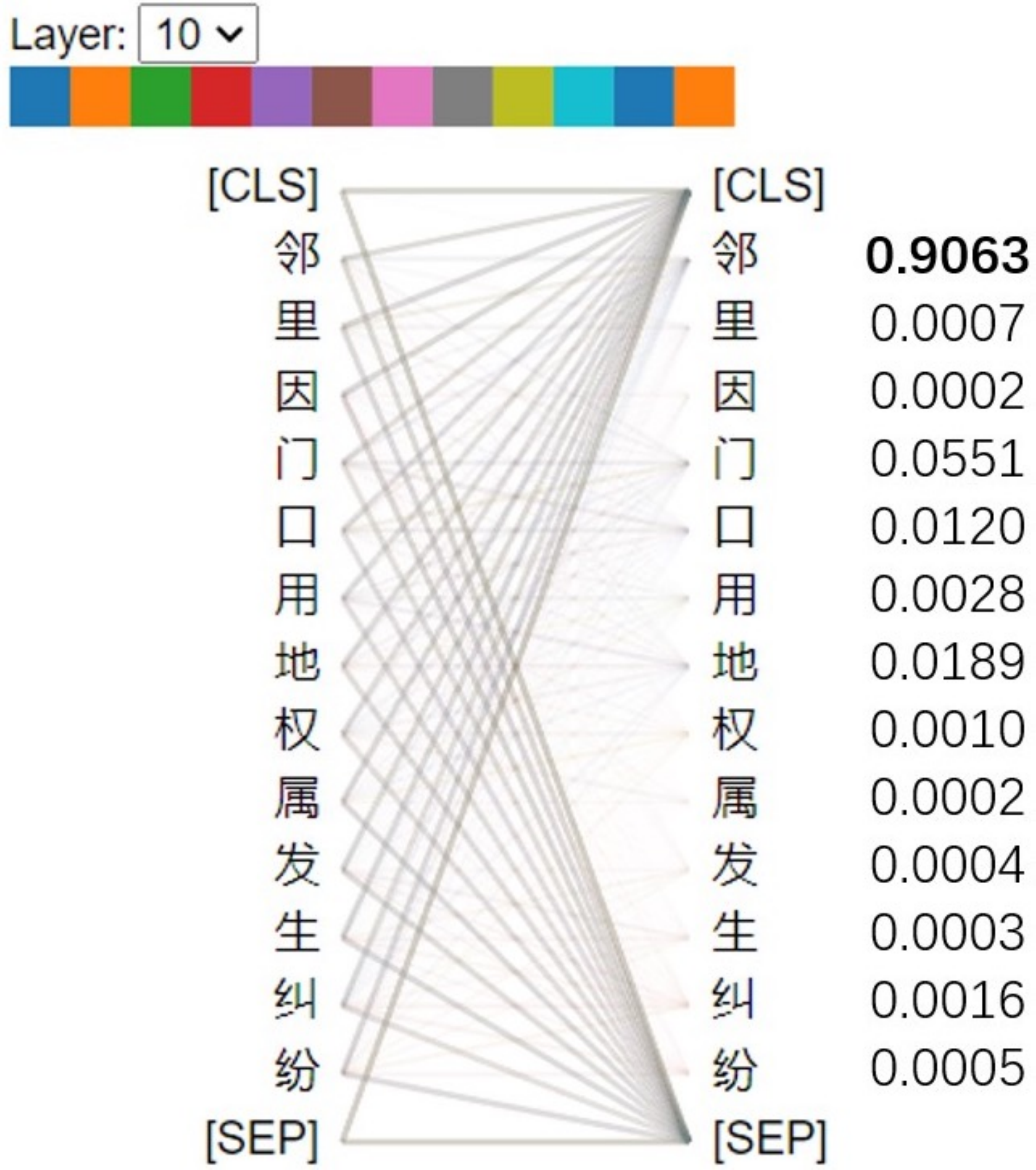}}
	\hspace{0.4in} 
	\subfigure[after enhancement]{
		\includegraphics[scale=0.3]{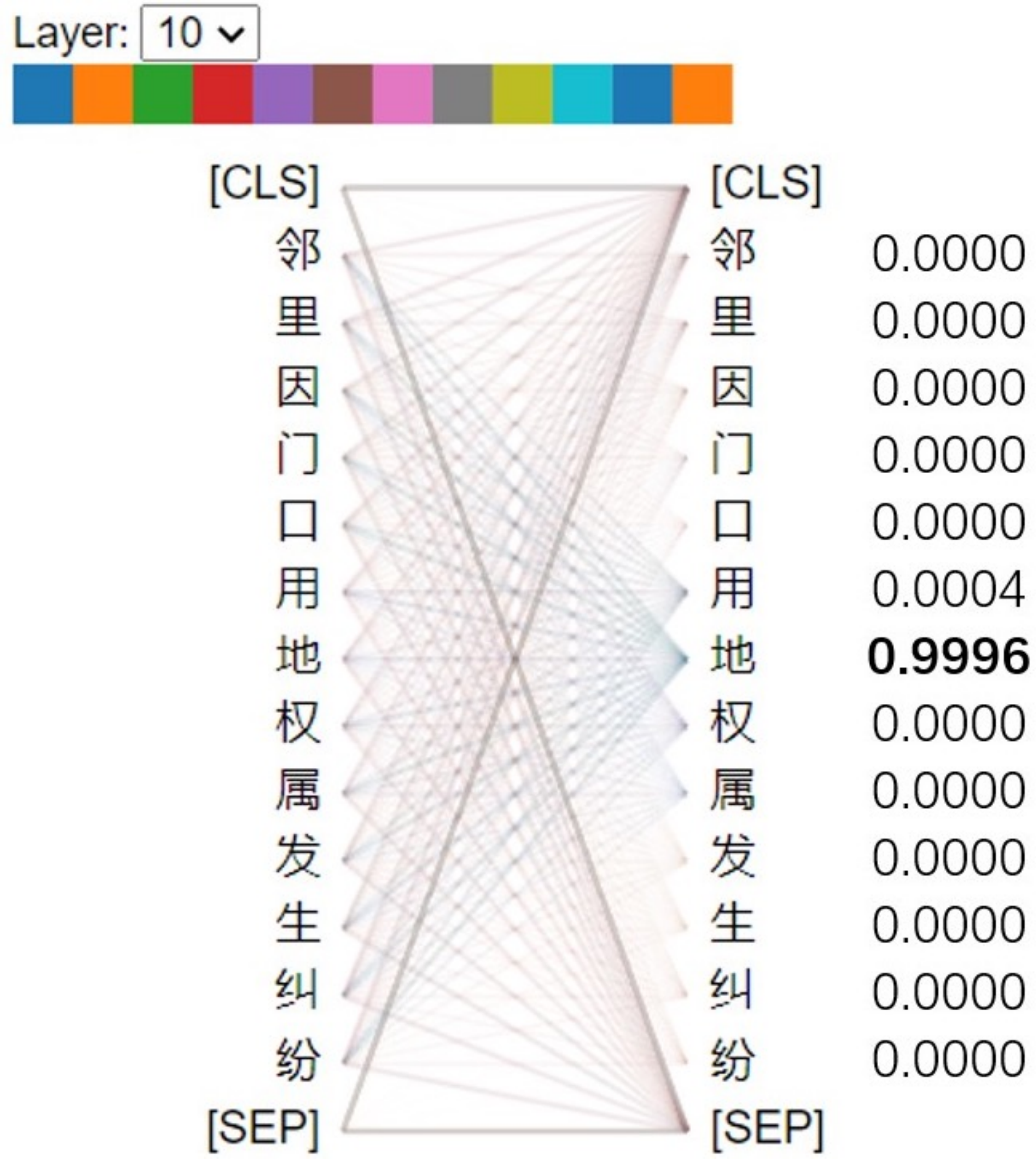}}
	}
	\captionsetup{font=scriptsize}\caption{The figure shows the change of the attention weights of the Transformer on the 10th layer before and after the enhancement of the model on sentences "中国足球彩票进球游戏2009年02月竞猜场次安排" (from the THUCNews dataset) and "邻里因门口用地权属发生纠纷" (from the social governance dataset). The darker the color of the line, the higher the attention weight the current token receives from other tokens. The right side of each picture shows the attention score of each token obtained by summing the attention weights from all tokens in the sentence, and then applying softmax. The maximum values are highlighted.}
\label{Figure-5}
\end{figure}

As shown in Table \ref{Table-4}, the attention mechanism from the enhanced model can extract the keywords more related to the topic than that from the original pre-trained language model.
In this section, we select two sentences from different datasets to show the shift in attention before and after enhancement.
Figure \ref{Figure-5} shows the shift in attention for the sentence "中国足球彩票进球游戏2009年02月竞猜场次安排(Chinese football lottery goal game February 2009 guess game schedule)" in the "Lottery" category of the THUCNews dataset and the sentence "邻里因门口用地权属发生纠纷(Neighborhood disputes over entrance land ownership rights)" in the "Land" category of the Social governance dataset.  
By comparing Figures 5(a) and (b), it can be seen that both of the model before and after enhancement have the largest attention on the token "彩(Lottery)", but the enhanced model has increased its attention to the category semantics, thus deepening its focus on the keywords most relevant to the category topic, and the attention weight has increased from 0.5473 to 0.9996.
Figures 5(c) and (d) show that, the original pre-trained model focuses most on the token "邻(Neighborhood)" in the sentence, but due to the training objectives of the model being to better differentiate the categories of the corpus, the attention was shifted more to the token "地(Land)" related to the "Land" category.
Specifically, the attention score of the token "地(Land)" increased from 0.0189 to 0.9996, and the score of the token "邻(Neighborhood)" decreased from 0.9063 to 0.
In addition, one commonality can be found in both experiments: the attention of the model before enhancement is more evenly distributed among each token of the sentence, but after the enhancement, the model almost entirely focuses on the key tokens related to the category, which undoubtedly enhances the effect of topic extraction.

\section{Conclusions}
In this paper, we propose an unsupervised text clustering and topic extraction framework ClusTop.
The sentence embeddings and the attention weights provided by the language model are used to tightly integrate two very related tasks, clustering and topic extraction.
The proposed framework doesn't require any manual annotation data and external knowledge, and it is flexible on the selection of pre-trained language models, dimensionality reduction and clustering algorithms, and has strong domain transferability.
The proposed framework consists of four parts: text representation based on the enhanced language model, dimensionality reduction, clustering and topic extraction based on the attention mechanism, among which the enhancement of the language model and the attention based topic extraction are the main contributions of our work.
The experimental results show that the enhanced language model performs better than the original pre-trained language model in all clustering evaluation metrics.
The attention based topic extraction method maximizes the function of attention mechanisms, and integrates topic extraction and clustering naturally, without stacking additional topic extraction algorithm afterwards.
The experimental results show that our attention based topic extraction method outperforms strong topic extraction baselines in the consistence with cluster topic semantics.
\par
As the proposed framework is independent of models, the overall clustering and topic extraction performance can be further improved with the continuous development of language models, dimension reduction and clustering algorithms.
We proposed a viable unsupervised approach to obtain the enhanced language model by training a classification model, but essentially, any approach that can make the language model more sensitive to cluster semantics can be used to replace the enhancement method in this paper.

\bibliographystyle{unsrt}
\bibliography{clusTop_v1}

\end{CJK}
\end{document}